\documentclass{article}

\usepackage[final, nonatbib]{neurips_2021}

\usepackage{microtype}
\usepackage{booktabs} 
\usepackage{enumitem}
\usepackage{hyperref}       
\hypersetup{
    colorlinks=true,
    citecolor = blue,
    linkcolor=blue
}
\usepackage{url}            
\usepackage{booktabs}       
\usepackage{amsfonts}       
\usepackage{nicefrac}       
\usepackage{microtype}      
\usepackage{outlines}
\usepackage{graphicx}
\usepackage{amsmath,bm}
\usepackage{color,soul}
\usepackage{amssymb}
\usepackage{algorithm}
\usepackage{algorithmic}
\usepackage{mathtools}
\usepackage{wrapfig}
\usepackage{centernot}
\usepackage{multirow}
\usepackage{multibib}
\usepackage{placeins} 
\usepackage{varwidth}
\usepackage{dsfont}
\usepackage{caption}
\usepackage{subcaption}
\usepackage{xcolor}

\usepackage[font=small,skip=2pt]{caption}

\usepackage{cleveref}

\newtheorem{assumption}{Assumption}

\newcommand\independent{\protect\mathpalette{\protect\independenT}{\perp}}
\def\independenT#1#2{\mathrel{\rlap{$#1#2$}\mkern2mu{#1#2}}}

\newcommand{\indep}{\raisebox{0.08em}{\rotatebox[origin=c]{90}{$\models$}}}

\title{MIRACLE: Causally-Aware Imputation via \\Learning Missing Data Mechanisms}

\author{
  Trent Kyono\thanks{Equal contribution} \\
  University of California, Los Angeles\\
  \texttt{tmkyono@ucla.edu} \\
  \And
  Yao Zhang$^*$ \\
  University of Cambridge \\
  \texttt{yz555@cam.ac.uk} \\
  \And
  Alexis Bellot \\
    University of Oxford \\
    Oxford, United Kingdom \\
    \texttt{alexis.bellot@eng.ox.ac.uk}\\
  \And
  Mihaela van der Schaar \\
  University of Cambridge \\
  University of California, Los Angeles\\
  The Alan Turing Institute \\
  \texttt{mv472@cam.ac.uk} \\  
}

\begin{document}

\maketitle

\begin{abstract}
Missing data is an important problem in machine learning practice. Starting from the premise that imputation methods should preserve the causal structure of the data, we develop a regularization scheme that encourages any baseline imputation method to be causally consistent with the underlying data generating mechanism. Our proposal is a causally-aware imputation algorithm (MIRACLE). MIRACLE iteratively refines the imputation of a baseline by simultaneously modeling the missingness generating mechanism, encouraging imputation to be consistent with the causal structure of the data. We conduct extensive experiments on synthetic and a variety of publicly available datasets to show that MIRACLE is able to consistently improve imputation over a variety of benchmark methods across all three missingness scenarios: at random, completely at random, and not at random.
\end{abstract}

\section{Introduction}
Missing data is an unavoidable byproduct of collecting data in most practical domains. In medicine, for example, doctors may choose to omit what they deem to be irrelevant information (e.g., some patients may be asked to get comprehensive blood tests while others don't), data may be explicitly omitted by the patient (e.g., avoiding questions on smoking status precisely because of their smoking habit) or simply misrecorded in electronic health systems (see e.g., \cite{bell2014handling,jakobsen2017and,sterne2009multiple}). 

\begin{minipage}{.65\textwidth}
Imputation algorithms can be used to estimate missing values based on data that was recorded, but their correctness depends on the type of missingness. For instance, expanding on the example above, younger patients may also be more likely to omit their smoking status. As illustrated in Figure \ref{m_graph}, the challenge is that implicitly conditioning inference on observed data introduces a spurious path of correlation between age and the prevalence of smoking that wouldn't exist with complete data. 
\end{minipage}
\hfill
\begin{minipage}{.33\textwidth}
\centering
\includegraphics[width=0.7\textwidth]{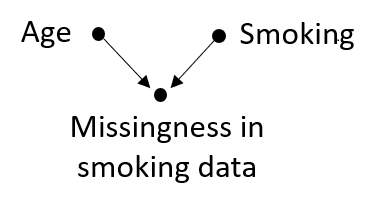}
\captionof{figure}{Missingness may introduce spurious dependencies.}
\label{m_graph}
\end{minipage}

Missing data creates a shift between the available missing data distribution and the target complete data distribution. It is a shift that may be explicitly modeled as missingness indicators in an underlying causal model (i.e., a missingness graph as proposed by Mohan et al. \cite{mohan-13-mgraphs}) as shown in Figure \ref{m_graph}. The learning problem is one of \textit{extrapolation}, learning with access to a missing data distribution for prediction and inference on the complete data distribution -- that is, generated from a model where all missingness indicators have been intervened on (interventions interpreted in the sense of Pearl \cite{pearl2009causality}) thus graphically removing the dependence between missingness and its causes, and any spurious correlations among its ancestors. 

With this causal interpretation, imputation of missing data on a given variable $Y$ from other observed variables $X$ is formulated as a problem of robust optimization,
\begin{align}
\label{robust_pop}
    \underset{\theta\in\Theta }{\text{minimize}}\hspace{0.1cm} \underset{P \in \mathcal P}{\sup}\hspace{0.1cm} \mathbb E_{(X,Y)\sim P} \left[(f_\theta(X)-Y)^2\right],
\end{align}
simultaneously optimizing over the set of distributions $\mathcal P$ arising from interventions on missingness indicators. Causal solutions -- i.e. imputation using functions of causal parents of each missing variable in the underlying causal graph -- are a closely-related version of this problem with an uncertainty set $\mathcal Q$ defined as any distribution arising from interventions on observed variables and variable indicators (see e.g. sections 3.2 and 3.3 in \cite{meinshausen2018causality}),
\begin{align}
\label{eq_intro_2}
    \underset{P \in \mathcal P}{\sup}\hspace{0.1cm} \mathbb E_{(X,Y)\sim P} \left[(f_\theta(X)-Y)^2\right] \leq \underset{P \in \mathcal Q}{\sup}\hspace{0.1cm} \mathbb E_{(X,Y)\sim P} \left[(f_\theta(X)-Y)^2\right],
\end{align}
since $\mathcal P \subset \mathcal Q$. Our premise is that causal solutions, i.e.  minimizing the right-hand-side of (\ref{eq_intro_2}), are expected to correct for spurious correlations introduced by distribution shift due to missing data and preserve the dependencies of the complete data for downstream analysis.

\subsection{Contributions}
In this paper, we propose to impute while preserving the causal structure of the data. Missing values in a given variable are replaced with their conditional expectation given the realization of its causal parents instead of the more common conditional expectation given all other observed variables, which absorbs spurious correlations. 

\begin{figure*}[t]
    \centering
    \includegraphics[width=1\linewidth]{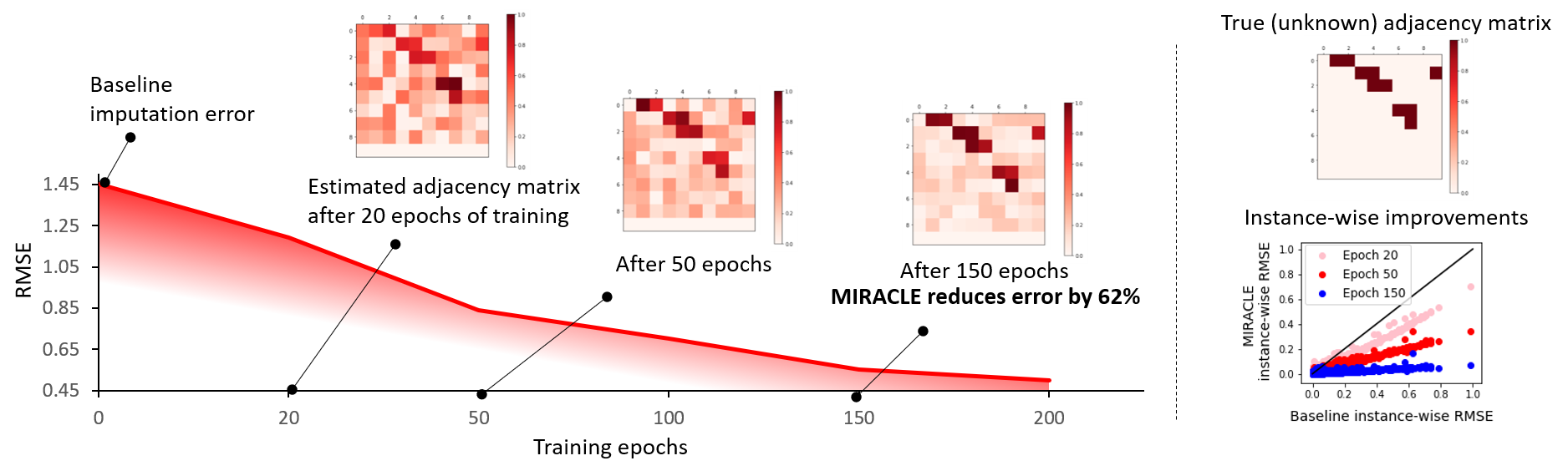}
    \caption{MIRACLE refines baseline imputation by simultaneously learning an $m$-graph using a bootstrap imputation loop that serves to incrementally regularize predictions with a learned causal graph. We plot average testing error and estimated causal graph as a function of training epochs on a synthetic data experiment described in 
    Section~\ref{section:experiments}. 
    The true causal structure (as an adjacency matrix) and imputation improvements for each missing value separately (each missing value with a corresponding dot) is shown in the right-most panel. 
    }
    \label{fig:intro}
\end{figure*}

We propose a novel imputation method called Missing data Imputation Refinement And Causal LEarning (MIRACLE). MIRACLE is a general framework for imputation that operates on any baseline (existing) imputation method. A visual description is given in Figure \ref{fig:intro}: given some initial imputation from a baseline method, MIRACLE refines its imputations iteratively by learning a missingness graph ($m$-graph) \cite{mohan-13-mgraphs} and regularizing the imputation function such that it is consistent with the causal graph generating the data, substantially improving performance. In experiments, we apply MIRACLE to improve six popular imputation methods as baselines. We present detailed simulations to demonstrate on synthetic and a variety of publicly available datasets from the UCI Machine Learning Repository \cite{uci} that MIRACLE can improve imputation in almost every scenario and never degrades performance across all imputation methods. 

\subsection{Related work}

The literature on imputation is large and varied. Still, most imputation algorithms work with the prior assumption that the missing data mechanism is ignorable, in the sense that imputation does not require explicitly modeling the distribution of missing values for imputation \cite{seaman2013meant}. Accordingly, classical imputation methods impute using a joint distribution over the incomplete data that is either explicit or implicitly defined through a set of conditional distributions. For example, explicit joint modeling methods include generative methods based on Generative Adversarial Networks \cite{GAIN,GAMIN,GAN_imp2016}, matrix completion methods \cite{MatrixCompletion}, and parametric models for the joint distribution of the data. Missing values are then imputed by drawing from their predictive distribution. The conditional modeling approach \cite{vanbuuren-MissingData} consists of specifying one model for each variable and iteratively imputing using estimated conditional distributions. Examples of discriminative methods are random forests \cite{MissForest}, autoencoders \cite{VAE-impute,DenoisingAE,MIWAE}, graph neural networks \cite{GRAPE}, distribution matching via optimal transport \cite{imputation-OT}, and multiple imputation using chained equations \cite{MICE1}. 

In a different line of research, Mohan et al., in a series of papers, see e.g. \cite{mohan-13-mgraphs,mohan2014testability}, explicitly considered missing data within the underlying causal mechanisms of the data. Subsequently, a range of related problems has been studied, including identifiability of distributions and causal effects in the presence of missing data, see e.g. \cite{shpitser-uai-2019,shpitser-uai-2015,shpitser-icml-2020},  testable implications relating to the causal structure using missing data \cite{mohan2014testability}, and causal discovery in the presence of missing data \cite{gain2018structure,tu2019causal}. Our focus, in contrast, is algorithmic in nature. We aim to develop an algorithm that improves imputation quality by leveraging causal insights represented as an estimated missingness graph learned from data.


\section{Background}\label{sect:back}


The basic semantic framework of our analysis rests on structural causal models (SCMs) (see e.g. Chapter 7 in \cite{pearl2009causality} for more details) explicitly introducing missingness indicators and their functional relationship with other variables, using in part the notation of \cite{mohan-13-mgraphs}. We define an SCM $\mathcal M$ as a tuple $(\bm  X, \bm R,\bm U,\mathcal F, P)$ where $\bm  X$ is a vector of $d$
endogenous variables and $\bm U$ is a vector of exogenous variables.\footnote{Essentially, $\bm X$ is the ground-truth features; $\bm U$ is the random noise in the data generating process.
} 
$\bm  R$ is the vector of missingness indicators that represent the status of missingness of the endogenous variables $\bm X$. Precisely, $R_j$ is responsible for the value of a proxy variable $Z_j$ of $X_j$, i.e., the observed version of $X_j$. For example, $Z_j$ is equal to $X_j$ if the corresponding record is observed ($R_j =1$), otherwise $Z_j$ is missing ($R_j =0$). $\mathcal F$ is a set of functions where each $f_X,f_R \in\mathcal F$ decide the values of an endogenous variable $X$ and a missingness indicator variable $R$, respectively. The function $f_X$ takes two separate arguments as parts of $\bm X$ (except $X$ itself) and $\bm U$, termed as  $\text{Pa}_X$ and $U_X$. That is, $X\leftarrow f_X(\text{Pa}_{X} , U_X )$ and  $R\leftarrow f_R(\text{Pa}_{R} , U_R )$.

The randomness in SCMs comes from the exogenous distribution $P_{\bm  U}(\bm u)$ where the exogenous variables in $\bm U$ are generated independently and are mutually independent. Naturally, through the functions in $\mathcal F$, the SCM $\mathcal M$ induces a joint distribution $P_{\bm X}(\bm x)$ over the endogenous variables $\bm X$, called the endogenous distribution. An intervention on some arbitrary random variables $\bm V$ in $\bm X$ and $\bm R$, denoted by $do(\bm  v)$, is an operation which sets the value of $\bm V$ to be $\bm v$, regardless of how they are ordinarily determined. For an SCM $\mathcal M$, let $\mathcal M_{\bm v}$ denote a submodel of $\mathcal M$ induced by intervention $do(\bm v)$. The interventional distribution $P_{\bm X}(\bm x|do(\bm v))$ induced by $do(\bm v)$ is defined as the distribution over $\bm X$ in the submodel $\mathcal M_{\bm v}$, namely, $P_{\bm X, \mathcal M_{\bm v}}(\bm x) = P_{\bm X}(\bm x|do(\bm v)) $. 

Each SCM in the context of missingness is associated with a $m$-graph $\mathcal G$ (e.g., Fig. 1a), which is a directed acyclic graph (DAG) where nodes represent endogenous variables $\bm X$ and missingness indicators $\bm R$, and arrows represent the arguments $\text{Pa}_{X}$ and $\text{Pa}_{R}$ of each function $f_X$ and $f_R$ respectively. By convention, exogenous variables $\bm U$ are often not shown explicitly in the graph.

\begin{figure}[!t]
    \centering
    \subfloat[A MCAR graph.]{\includegraphics[width=2.7cm]{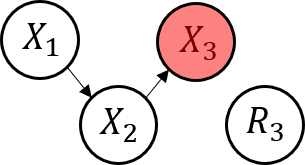} }%
    \quad
    \subfloat[A MAR graph.]{\includegraphics[width=2.7cm]{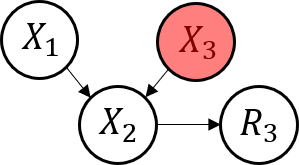} }
    \quad
    \subfloat[A MNAR graph.]{\includegraphics[width=3.3cm]{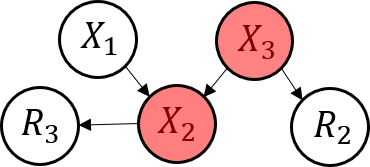} }
    \quad
    \subfloat[Self-masking missingness.]{\includegraphics[width=2.1cm]{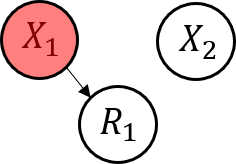} }
    \caption{Example graphs. $\bm X = (X_1, X_2, X_3)$ are endogenous variables and $\bm R = (R_{1},R_{2}, R_{3})$ are missing data indicators. Red shaded variables are not always observed while white shaded variables are always observed.} 
    \label{fig:miss_graphs}
    \vspace{-1em}
\end{figure}

\begin{assumption}[Missingness indicators are not causes] \label{as:1}
    No missingness indicator in $\bm R$ can be the cause of the endogenous variables $\bm X$, i.e., the arguments of the functions generating $\bm X$.
\end{assumption}
\begin{assumption}[Causal sufficiency] \label{as:2}
    Exogeneous variables $\bm U$ are mutually independent, i.e., all common parents of the endogenous variables are included in $\bm X$.
\end{assumption}
\begin{assumption}[No self-masking missingness] \label{as:3}
    Self-masking missingness refers to missingness in a variable that is caused by itself. In the $m$-graph this is depicted by an edge from $X_j$ to $R_j$ (as shown in Figure \ref{fig:miss_graphs} (d)). We assume that there is no such edges in the $m$-graph.
\end{assumption}
\begin{assumption}[Observed root nodes] \label{as:4}
    The endogenous variables $X_j$ such that $\text{Pa}_{X_j} = \emptyset$ (i.e., the root nodes) in the $m$-graph are always observed ($R_j=1$ with probability 1).
\end{assumption}

We make the four assumptions above throughout the following sections.
Assumption \ref{as:1} and \ref{as:2} are employed in most related works using $m$-graphs (see e.g. \cite{mohan2014testability,mohan-13-mgraphs}). Assumption~\ref{as:1} is valid, for example, if $\bm R$ is generated in the data collection process after the variable values are assigned. Consequently, under this assumption, if two endogenous variables of interest $X_1$ and $X_2$ are not $d$-separated by some variable $ X_3$, they are not $d$-separated by $ X_3$ together with their missingness indicators $R_1$ and $R_2$.
We denote an independent relation in a data distribution by ``$\indep$" and $d$-separation in a $m$-graph by ``$\indep_d$". We assume the data distribution is faithful to a $m$-graph, meaning that the two independencies are equivalent. As shown in Figure \ref{fig:miss_graphs}, data is missing completely at random (MCAR) if $\bm X \indep_{d} \bm R$ holds in the $m$-graph, missing at random (MAR) if for any endogenous variable $X_j$, $R_j \indep_d X_j \mid \bm X_{-j}$ holds, and missing not at random (MNAR) otherwise, as stated in \cite{mohan-13-mgraphs}. 
If Assumption~\ref{as:3} is violated, we are unable to learn the missingness for self-masked variables. 
Assumption~\ref{as:4} is necessary for imputing all the missing variables from their causal parents.
These assumptions are imperative for MIRACLE to provide improved imputations by leveraging the causal structure of the underlying data generating process. In our experiments (Section~\ref{section:experiments}), we apply MIRACLE to real-world datasets where these assumptions are not guaranteed.

\subsection{Why is imputation prone to bias?}\label{subsect:problem}

The reason for considering the causal structure of the underlying system is that when learning an imputation model from observed data, implicitly conditioning on some missingness indicators in $\bm R$  induces spurious dependencies that would not otherwise exist. For example, in a graph $X_1 \rightarrow R_3 \leftarrow X_2$, conditioning on $R_3=1$ induces a dependence between $X_1$ and $X_2$. In general, the distributions $P_{\bm X}(\bm x | \bm R = \bm r)$ and $P_{\bm X}(\bm x|do(\bm r) )$ differ unless missingness occurs completely at random, and motivates an interpretation of the problem as domain generalization, training on data from one distribution ultimately to be applied on data from a different distribution that, in our case, arises from missing data (i.e., interventions on missingness indicators). This shift is not addressed in the imputation methods that only use the feature correlations.

\section{MIRACLE} 
In this section, we propose to correct for the shift in distribution due to missing data by searching for causal solutions and explicitly refining imputation methods using a penalty on the induced causal structure. 
In practice, we have $n$ $i.i.d.$ realizations of the observed version of ${\bm X} \in \mathbb{R}^d$, concatenated as a incomplete data matrix $\mathbf{X}\in\mathbb R^{n\times d}$, together with missingness indicators concatenated in a matrix $\mathbf R \in \{0,1\}^{n\times d}$. We use here the same bold uppercase notation for sets of variables and matrices of observations but their meaning should be clear from the context. Our goal is to impute the unobserved values in $\mathbf X$ using each variable's causal parents. We define the \textit{imputed} data $\mathbf{\tilde X}\in\mathbb R^{n\times d}$,
\begin{align*}
    \mathbf{\tilde X}= \mathbf R \odot \mathbf{X} + (1 - \mathbf R) \odot \mathbf{\hat X}
\end{align*}
where $\odot$ is the element-wise product of matrices and $\mathbf{\hat X}$ is an estimate of the complete data matrix.

\subsection{Network architecture}
In this section, we describe our approach for estimating $\mathbf{\hat X}$. 
Let $d_S\leq d$ be the number of partially observed features, i.e., missing for at least one realization. $S$ is the set of missing features indices. The imputation network is defined as a function $f: \mathbb{R}^d \rightarrow  \mathbb{R}^{d}  \times [0,1]^{d_S} $ that takes an initially imputed dataset $\tilde{\mathbf{X}}^{(0)}$ (using an existing baseline imputation method), and returns two quantities:
\begin{enumerate}
    \item A refined imputation $\mathbf{\hat{X}}$.
    \item An estimation of the probabilities of features $X_{ij}$ being missing, $i=1...,n$ and $j\in S$ . 
\end{enumerate}

\begin{figure*}[t]
    \centering
    \vspace{-0.5cm}
    \includegraphics[width=\linewidth]{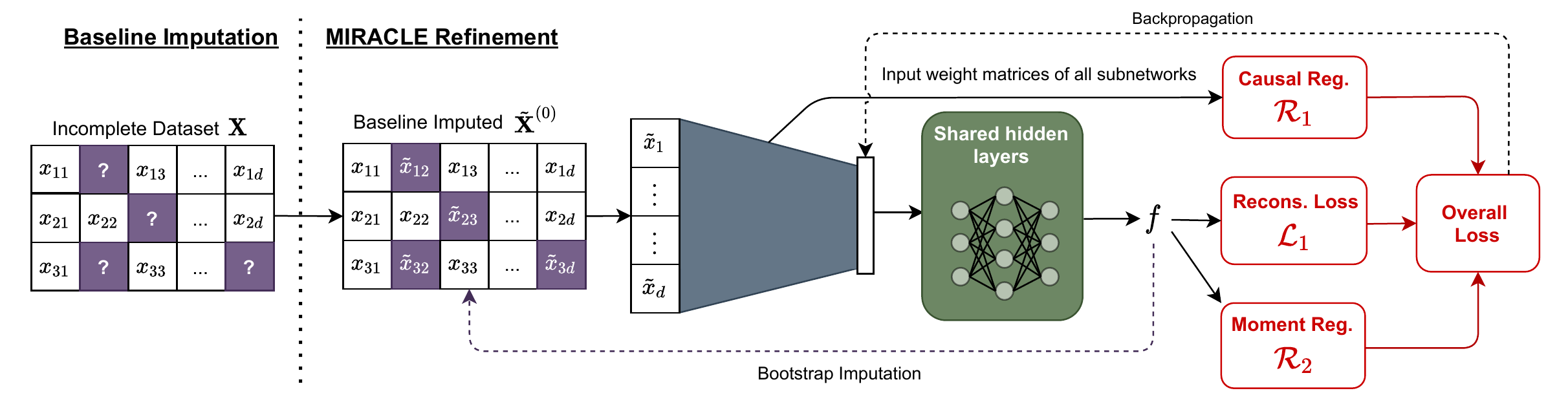}
    \caption{Network and optimization diagram for MIRACLE.
    }
    \vspace{-0.5cm}
    \label{fig:schematic}
\end{figure*}

A depiction of the network architecture and optimization algorithm is shown in Figure \ref{fig:schematic}. The architecture is constructed with respect to the assumptions shown in Section \ref{sect:back}.
Our model $f$ is decomposed into two sub-networks, $f = (f^{(imp)}, f^{(miss)})$, responsible for imputing the unobserved data and estimate the probabilities of missingness, respectively. The imputation network has $d$ components, $f^{(imp)} = (f^{(imp)}_1,\dots,f^{(imp)}_d)$, one for each variable, and the missingness network has $d_S$ components, $f^{(miss)} = (f^{(miss)}_1,\dots,f^{(miss)}_{d_S})$. Each component, for both networks, has separate input and output layers but shared hidden layers (of size $h$). Let $\mathbf W^{(imp)}_{1,j}$ and $\mathbf W^{(miss)}_{1,j}$ denote the $h \times d$ weight matrix (we omit biases for clarity) in the input layer of $f_j^{(imp)}$ and $f_j^{(miss)}$ respectively. The $j$-th column of
$\mathbf W^{(imp)}_{1,j}$ and $\mathbf W^{(miss)}_{1,j}$ is set to $\bm 0$.
Let $\mathbf W_m\in\mathbb R^{h\times h}$, for $m = 2, \dots , M - 1$, denote the weight matrix of each hidden layer and let $\mathbf W^{(imp)}_{M,j}$ and $\mathbf W^{(miss)}_{M,j}$, be the $1 \times h$ dimensional output layers of each sub-network. 
The imputation network prediction is given by,
\begin{equation*}
\label{subnetwork}
 f^{(imp)}_j(\mathbf{x}) := \mathbf W^{(imp)}_{M,j}\phi(\cdots\phi(\mathbf W_2\phi( \mathbf W^{(imp)}_{1,j}\mathbf{x}    )) ),
\end{equation*}
for $j=1,\dots,d$. And similarly, the missingness network prediction is given by,
\begin{equation*}
\label{subnetwork2}
 f^{(miss)}_j(\mathbf{x}) := \sigma\big(\mathbf W^{(miss)}_{M,j}\phi(\cdots\phi(\mathbf W_2\phi( \mathbf W^{(miss)}_{1,j}\mathbf{x}    )) )\big),
\end{equation*}
for $j=1,\dots,d_S$, where $\phi(\cdot)$ is the ELU activation function and $\sigma$ is the sigmoid function. 
Our network is optimized with respect to three objectives. First, to accurately predict missing values, second, to faithfully encode the causal relationships given by the underlying $m$-graph, and third to satisfy a moment constraint of the missing data mechanism on the imputed values.

\subsection{Reconstruction loss} 

The first objective is to train $f$ to correctly reconstruct each feature from the observed data using a reconstruction loss,
\begin{equation*}
\label{eqi:obj_l}
   \mathcal L_1 = \frac{1}{n}\Bigg(\sum_{i=1}^{n}\big\| \mathbf{x}_i \odot \mathbf r_i - f^{(imp)}(\tilde{\mathbf{x}}_i^{(0)})\odot \mathbf r_i\big\|^2 + \sum_{i=1}^{n}\mathrm{CrossEntropy}\Big[\tilde{\mathbf{ r}
   }_i,f^{(miss)}(\tilde{\mathbf{x}}_i^{(0)})  \Big] \Bigg),
\end{equation*}
where $\mathbf{x}_i$ and $\tilde{\mathbf{x}}_i^{(0)}$ are the realized and imputed feature vector of the $i$-th instance, $\tilde{\mathbf{ r}
   }_i$ are the $d_S$ components of $\mathbf{r}_i$ that are missing for at least one instance. The first loss term  is for reconstructing the observed features, and the second loss term is for estimating the probabilities of missingness.

\subsection{Causal regularizer}
The second objective is to ensure that the dependencies defined by $f$ correspond to a DAG over the features $X$ and the missing indicators $R_S$, which enforces that the learned functional dependencies recover a DAG in the equivalence class of causal graphs over the observed data. 
Enforcing the acyclicity of the dependencies induced by a continuous function $f$ is originally proposed in \cite{zheng2018notears,zheng2019sparseDAGS}. Define a binary adjacency matrix $\mathbf B\in\mathbb \{0,1\}^{(d+d_S)\times (d+d_S)}$; $[\mathbf  B]_{k,j}=0$ (i.e., the $l_2$-norm of the $k$-th column of the matrix $\mathbf W^{(imp)}_{1,j}$ or $\mathbf W^{(miss)}_{1,j}$ is $0$) is a realistic and sufficient condition for achieving $\partial_k f_{j}=0$. The adjacency matrix $\mathbf B$ of the graph induced by the learned $f$ is acyclic if and only if,
\begin{align}
   \mathcal{R}_1 = \tfrac{1}{2}h^2(\mathbf B)+h(\mathbf B),
\end{align}
is equal to zero, where $h(B):= \operatorname{Tr}(\exp\{\mathbf  B \odot \mathbf  B\})-(d+d_S)$ and $\exp(\cdot)$ is the matrix exponential. 

\textbf{Remark 1.} Existing imputation methods based on feature correlations essentially assume an undirected (non-causal) graph between the features. Further, acyclicity is a realistic and practical assumption to make on the static datasets collected by human experts. In nature, most data distributions generate their features in some order. In a directed graph, a cycle means a path starts and ends at the same node. This is unlikely to happen in the data generating process if not considering variables over time, i.e., time-series data. By enforcing acyclicity, MIRACLE only uses the causal parents for imputation, which is less biased by spurious dependencies that only exist in the observed data.





\subsection{Moment regularizer}

The third objective leverages a set of moment constraints in the missingness pattern to improve imputation. 
Assume $\xi_j  = P \big( R_j =1 \mid \text{Pa}_{R_j})\big)\in (\delta ,1-\delta )$, for some $\delta>0$. The following derivation holds for MAR or MCAR missingness patterns only. It holds that,
\begin{equation}
\label{equ:exp}
\begin{split}
\mathbb E \left\{ \frac{R_j X_j}{\xi_j}  \right\}
    =\mathbb E \left\{   \mathbb E \left[ \frac{R_j X_j}{\xi_j}\ \big| \  X_j, \text{Pa}_{R_j} \right] \right\} 
    = \mathbb E \left\{   \frac{X_j }{\xi_j}  \mathbb E \left[   R_j \mid X_j, \text{Pa}_{R_j} \right] \right\}
    = \mathbb E \left\{ X_j \right\},
   \end{split}  
\end{equation}
where the third equality follows from the MAR assumption ($R_j \independent X_j \mid \bm X_{-j}$). Under the MCAR assumption, this derivation holds trivially since in that case $R_j \independent X_j$.

We can use the missingness and imputation networks to enforce the above equality algorithmically, ensuring the left hand side equals the right hand side in the empirical version of \eqref{equ:exp} as follows,
\begin{equation*}
\begin{split}
\mathcal{R}_2  = \sum_{j=1}^{d_S} \left[ \hat{\tau}_{j,\text{SIPW}} -  \hat{\tau}_{j,\text{mean}}       \right]^2 = 
\sum_{j=1}^{d_S} \left[
\left(\sum_{i=1}^n e_{ij}  r_{ij}\right)^{-1} \sum_{i=1}^n e_{ij} r_{ij}x_{ij}   - \frac{1}{n}\sum_{i=1}^{n}f_{S[j]}^{(imp)}(\tilde{\mathbf x}_i^{(0)})   \right]^2,
\end{split}
\end{equation*}
where $e_{ij} = 1/f_j^{(miss)}(\tilde{\mathbf x}_i^{(0)})$, and $S[j]$ is the $j$-th element of $S$, i.e., the index of the $j$-th missing feature. Minimizing $\mathcal{R}_2$  forces the two estimators of $\mathbb E \left\{ X_j \right\}$  to match, the stabilized inverse propensity score weighting (SIPW) estimator $ \hat{\tau}_{j,\text{SIPW}}$ \cite{robins2007comment} using the missingness network $f_j^{(miss)}$ (in $e_{ij}$) and the mean estimator $\hat{\tau}_{j,\text{mean}} $ using the imputation network  $f_{S[j]}^{(imp)}$.

\textbf{Remark 2.}  We hypothesize this mechanism can improve performance for two reasons. First, the missing data mechanism $P(R_j=1\mid \text{Pa}_{R_j})$ can be a simpler function that takes less samples to learn than the function that generates the feature $j$, $\mathbb E\big[X_j|\text{Pa}_{X_j}\big]$. Then the SIPW estimator based on $f_j^{(miss)}$ will converge to the true mean faster than the estimator based on $f_{S[j]}^{(imp)}$. Second, in $\mathcal{R}_2$, the mean estimator using $f_{S[j]}^{(imp)}$ is based on all the samples; $f_{S[j]}^{(imp)}$ is trained to produce predictions on the samples with missing feature $j$ for the sake of matching the SIPW estimator. By contrast, without the regularizer $\mathcal{R}_2$, $f_{S[j]}^{(imp)}$ is solely trained on the samples with observed feature $j$, and its performance may fail to generalize to data with missing feature $j$.

\subsection{Bootstrap Imputation}

Discovering a causal graph requires complete data.  However, this is not the case for missing data problems.  Because of this, we require that MIRACLE be seeded by another imputation method.  Imputed values are iteratively refined by MIRACLE, hence ``bootstrapping'', to potentially converge to a new imputation that minimizes MIRACLE's objective (including causal and moment regularizers).
MIRACLE's objective for optimization is, 
\begin{align}\label{eq:overall_loss}
  \mathcal{L} = \mathcal{L}_1 + \beta_1\mathcal{R}_1 +  \beta_2\mathcal{R}_2, 
\end{align}
where $\beta_1$ and $\beta_2$ are hyperparameters that define the strength of regularization.
We iteratively update the baseline matrix $\tilde{\mathbf{\bm X}}^{(0)}$ with a new imputed matrix $\tilde{\mathbf{\bm X}}$ given by MIRACLE every ten epochs in training.
With increasing epochs, stochastic optimization minimizes the loss for the imputed matrices that respect the causal and moment regularization.
In theory, this is analogous to supervised training a denoising autoencoder (DAE) \cite{Vincent2010StackedDA,DAE2013, DAE2018}, but differs only by the fact that ``noise'' comes from prior or previous imputations.  In training DAE, the input samples are corrupted by independent noise with each epoch, yet convergence is still guaranteed \cite{arora2014provable}. 
In our experiments, we demonstrate that bootstrap imputation indeed converges on multiple datasets and baseline methods. In Appendix~\ref{appdx:complexity}, we provide experiments on the computational complexity of MIRACLE.




\begin{figure}[t]
    \centering
    \vspace{-0.05cm}
    \begin{subfigure}[b]{0.49\textwidth}
         \centering
         \includegraphics[width=\textwidth]{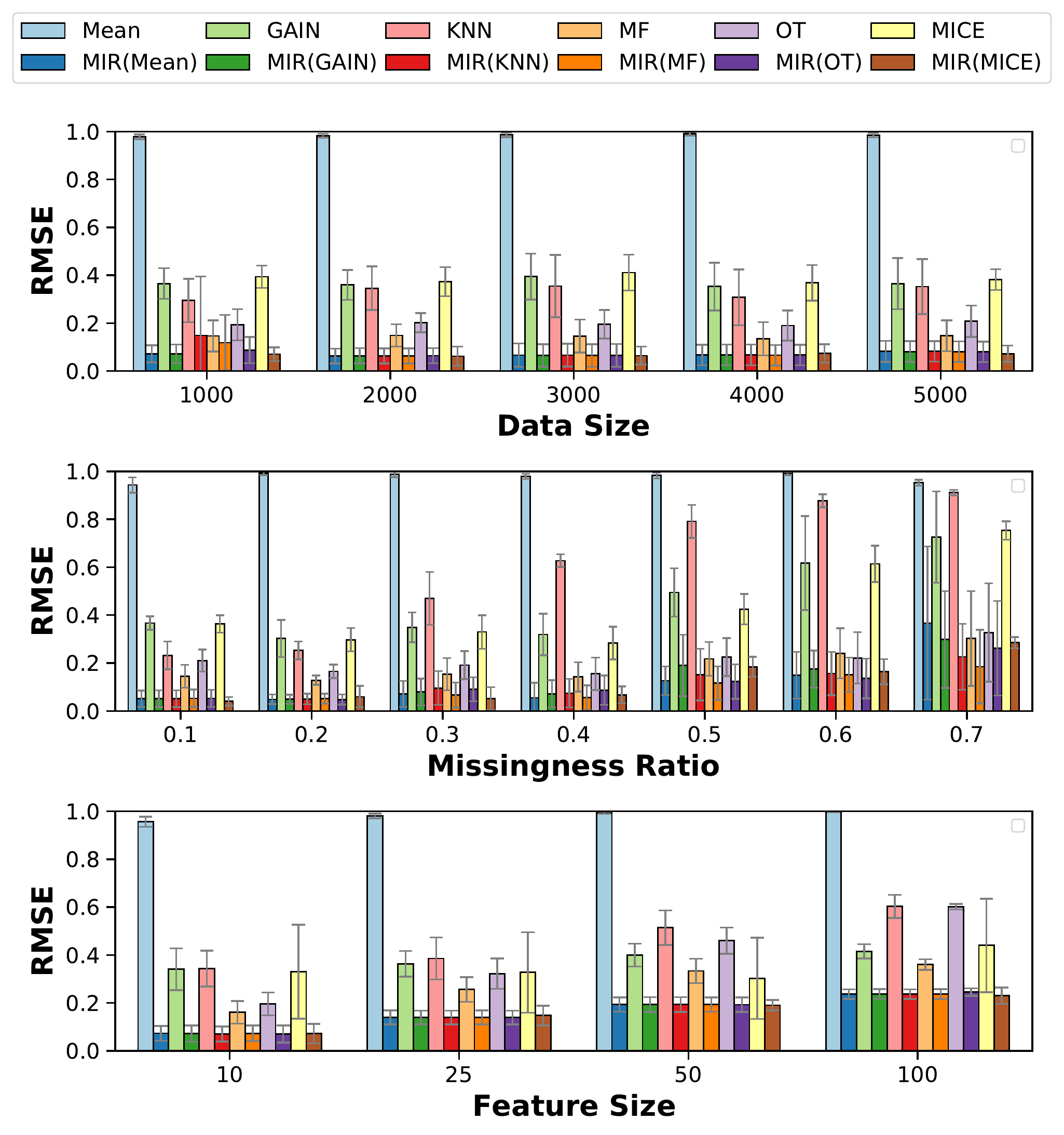}
         \caption{MAR}
         \label{fig:y equals x}
     \end{subfigure}
     \hfill
     \begin{subfigure}[b]{0.49\textwidth}
         \centering
         \includegraphics[width=\textwidth]{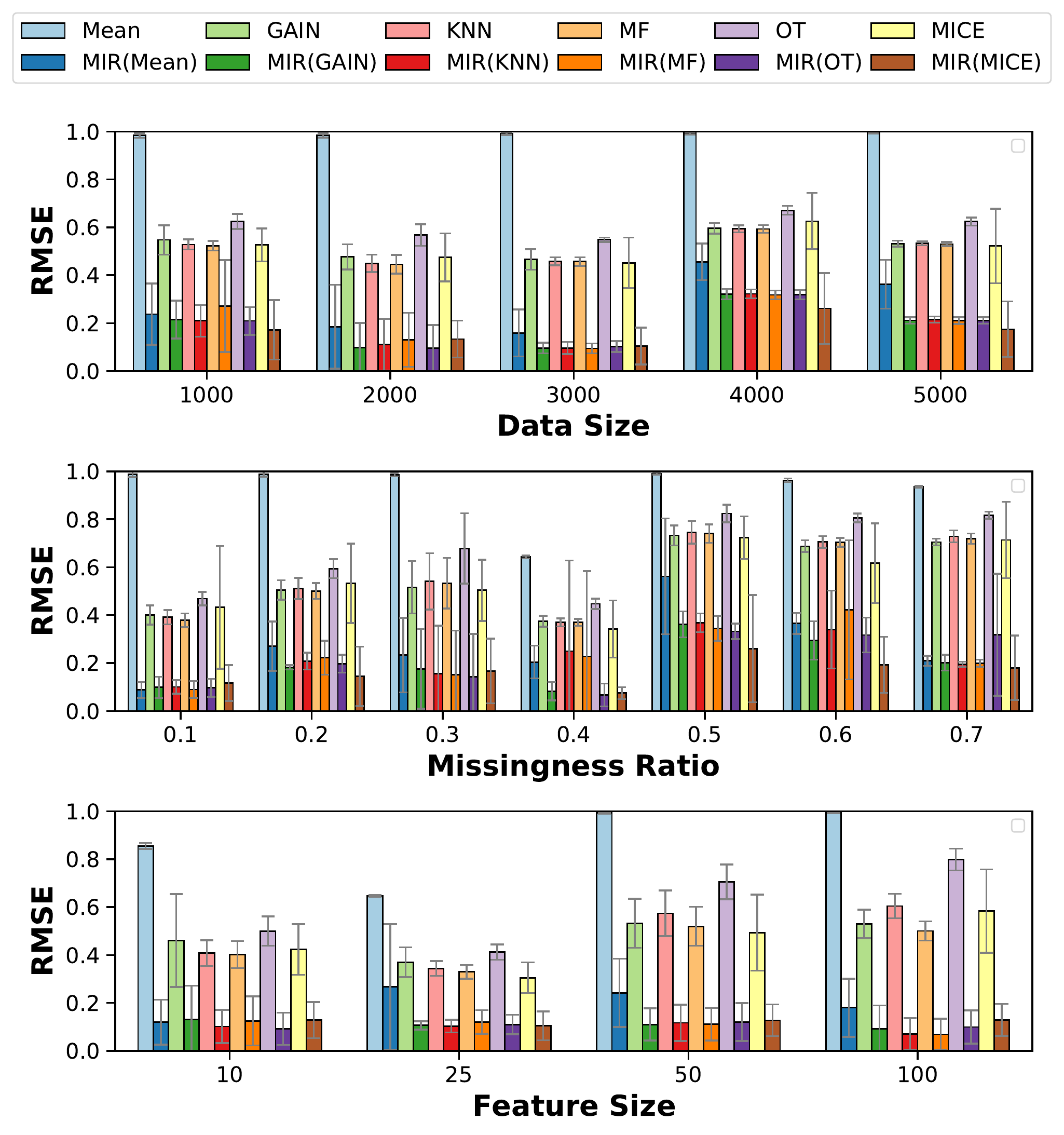}
         \caption{MNAR}
         \label{fig:three sin x}
     \end{subfigure}
     \hfill

    \vspace{-0.cm}
    \caption{\textbf{Experiments on MAR (left) and MNAR (right) synthetic data} in terms of RMSE over varying \emph{dataset sizes} \textbf{(top)}, \emph{missingness rates} \textbf{(middle)}, and \emph{feature sizes} \textbf{(bottom)}.  Note that we show the average error over a variety of DAG instantiations and target variables, thus the magnitude and standard deviation of errors vary significantly between runs. See Appendix~\ref{appdx:additionalsynth} for MCAR results.}
    \label{fig:synth_MAR}
    \vspace{-0.10cm}
\end{figure}

\section{Experiments} \label{section:experiments}

In this section, we validate the performance of MIRACLE using both synthetic and a variety of real-world datasets.  
\begin{enumerate}
    \item In the first set of experiments, we quantitatively analyze MIRACLE on synthetic data generated from a known causal structure.
    \item In the second set of experiments, we quantitatively evaluate the imputation performance of MIRACLE using various publicly available UCI datasets \cite{uci}.
\end{enumerate}  

\textbf{General set-up.} We conduct each experiment five times under random instantiations of missingness.  We report the RMSE along with standard deviation across each of the five experiments. Unless otherwise specified, missingness is applied at a rate of 30\% per feature.  For MCAR, this is applied uniformly across all features. For MAR, we randomly select 30\% of the features to have missingness caused by another disjoint and randomly chosen set of features.  Similarly, we randomly select 30\% of features to be MNAR.  We induce MAR and MNAR missingness using the methods outlined in \cite{GAIN}, and we provide more details in Appendix~\ref{appx:missingness}.  

We use an 80-20 train-test split.  We performed a hyperparameter sweep (log-based) for $\beta_1$ and  $\beta_2$ with ranges between 1e-3 and 100.  By default we have  $\beta_1$ and  $\beta_2$ set to 0.1 and 1, respectively. 

\textbf{Evaluating imputation.}
For each subsection below, we present three model evaluations in terms of missingness imputation performance, label prediction performance of a prediction algorithm trained on imputed data and the congeniality of imputation models.

\begin{itemize}
    \item \textbf{Missingness imputation performance} is evaluated with the root mean squared error comparing the imputed missing values with their actual unobserved values.
    \item \textbf{Label prediction performance} of an imputation model is its ability to improve the post-imputation prediction.  By post-imputation, we refer to using the imputed data to perform a downstream prediction task.  To be fair to all benchmark methods, we use the same model (support vector regression) in all cases. 
    \item \textbf{The congeniality} of an imputation model is its ability to impute values that respect the feature-label relationship post imputation.  Specifically, we compare, support vector parameters, $w$,  learned from the complete dataset with the parameters $\hat{w}$, learned from the imputed dataset.  
    We report root mean square error $(||w - \hat w ||^2)^{1/2}$ for each method. Lower values imply better congeniality \cite{GAIN}.
\end{itemize}

\paragraph{Baseline imputation methods.}
We apply MIRACLE imputation over a variety of six commonly used imputation baseline methods: (1) mean imputation using the feature-wise mean, (2) a deep generative adversarial network for imputation using GAIN \cite{GAIN} (3) $k$-nearest neighbors (KNN) \cite{knn-impute} using the Euclidean distance as a distance metric of each missing sample to observed samples, (4) a tree-based algorithm using MissForest (MF) \cite{MissForest}, (5) a deep neural distribution matching method based on optimal transport (OT) \cite{imputation-OT}, and (6) Multivariate Imputation by Chained Equations (MICE) \cite{MICE1}.  For each of the baseline imputation methods with tunable hyperparameters, we used the published values. We implement MIRACLE using the \texttt{tensorflow}\footnote{Source code at \url{https://github.com/vanderschaarlab/MIRACLE}.} library. Complete training details and hyperparameters are provided in Appendix~\ref{appx:training_details}.

\textbf{Additional experiments.} In Appendix~\ref{appdx:additionalsynth} and ~\ref{appx:additional_real_dataset}, we also evaluate MIRACLE in terms of providing imputations that improve \textbf{predictive performance} \cite{GAIN, GRAPE}, as well as, in its ability to impute values that respect the feature-label relationship, i.e., \textbf{congeniality} \cite{congeniality1, congeniality2}. We also investigate the impact of the graphical missingness location on imputation performance on synthetic data in Appendix~\ref{apx:missingness}.

\subsection{Synthetic data}
In this subsection, we evaluate MIRACLE on synthetic data.  In doing so, we can control aspects of the underlying data generating process and possess oracle knowledge of the DAG structure.

\textbf{Data generating process.}
We generate random Erdos-Renyi graphs with functional relationships from parent to children nodes. At each node, we add Gaussian noise with mean 0 and variance 1.  For a complete description of the underlying data generating process, see Appendix~\ref{appx:synth_DGP}.

\textbf{Synthetic results.}
In Figure~\ref{fig:synth_MAR}, we show experiments of MIRACLE on synthetic MAR data in terms of RMSE. Our experiments show that MIRACLE is able to significantly improve imputation over each of the baseline imputation methods. 
Figure~\ref{fig:synth_MAR} shows MIRACLE improves performance over each baseline method across various \textbf{dataset sizes, missingness ratios, and feature sizes (DAG sizes)}.  We show results for MCAR in Appendix~\ref{appdx:additionalsynth}.


\subsection{Experiments on real data}

\begin{figure*}[!h]
    \centering
    \vspace{-0.2cm}
    \includegraphics[width=1\linewidth]{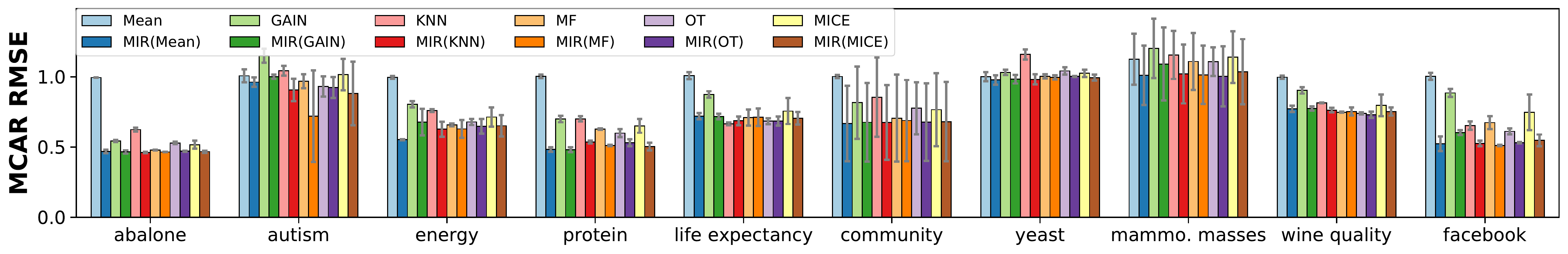} 
    \includegraphics[width=1\linewidth]{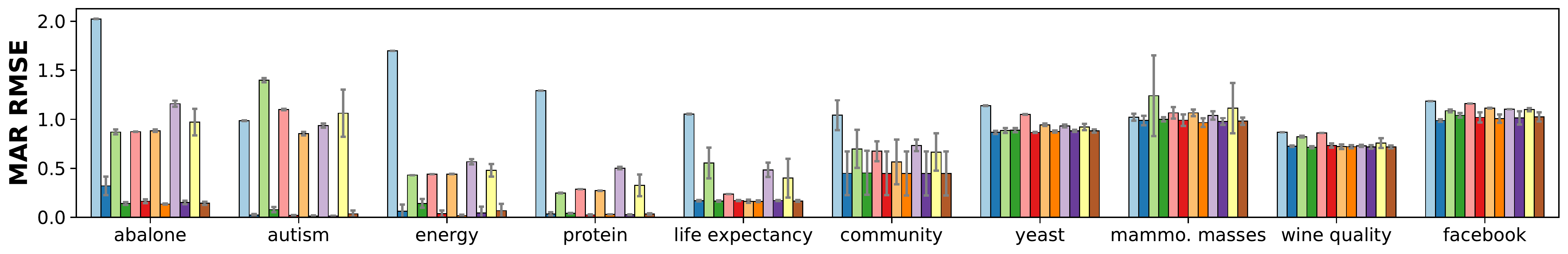} 
    \includegraphics[width=1\linewidth]{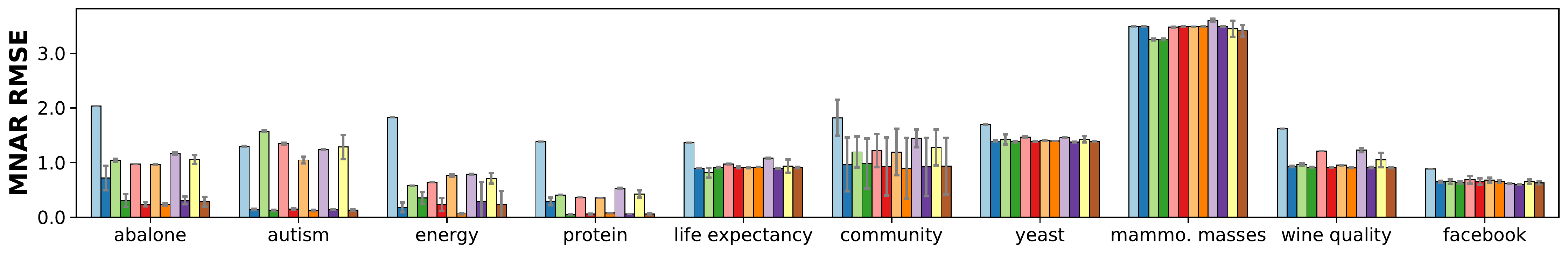} 
             \vspace{-0.45cm}
    \caption{\textbf{MIRACLE on real data}.  MIRACLE improves all baselines across MCAR \textbf{(top)}, MAR \textbf{(middle)}, and MNAR \textbf{(bottom)}. 
    Results for predictive error and congeniality can be found in the Appendix~\ref{appx:additional_real_dataset}.}
    \label{fig:real_uci}
    \vspace{-0.05cm}
\end{figure*}

In Figure~\ref{fig:real_uci} we show experiments of MIRACLE on real data.
We perform experiments on several UCI datasets used in \cite{GAIN,GRAPE,imputation-OT}: Autism, Life expectancy, Energy, Abalone, Protein Structure, Communities and Crime, Yeast, Mammographic Masses, Wine Quality, and Facebook Metrics. In Figure~\ref{fig:real_uci}, the improvements of MIRACLE are minimal for MCAR (except for mean imputation). This agrees with our discussion in Section~\ref{subsect:problem}, because the baseline imputations are not biased in the MCAR setting where $\bm X \indep_{d} \bm R$ holds in the $m$-graph.  Conversely for the MAR and MNAR settings, as expected, we observe MIRACLE  has an significant improvement on some of the datasets, such as Abalone, Autism, Energy and Protein Structure.  As discussed in Section \ref{sect:back}, MIRACLE can improve the baseline imputation under Assumptions \ref{as:1}-\ref{as:4}, which may not hold in these real-world datasets. Nevertheless, we observe that MIRACLE never degrades performance relative to its baseline imputation on any dataset. Furthermore, no baseline imputer is optimal across the datasets.  In almost all cases, applying MIRACLE to any baseline results in the lowest error. We show a similar gain for this experiment with MCAR and MNAR in Appendix~\ref{appx:additional_real_dataset}.
An \textbf{ablation study} on our overall loss is provided in Appendix~\ref{appx:ablation}.


\subsection{Causal discovery and imputation performance}
In our experiments, we observe a positive correlation between the quality of learned DAGs (and causal parents) with imputation performance.
Consider the left-most plot in Figure~\ref{fig:improvements} using OT as a baseline imputer under MAR on our real data sets.
Here, we do not have oracle knowledge but assume that the sparseness of the learned DAG implies a coherent DAG.
We observe that MIRACLE has the most performance gain when fewer causal parents are identified for the missing variable in the learned DAGs. When MIRACLE is less able to isolate causal parents for prediction, the learned DAGs contains many spurious edges, and MIRACLE only has marginal improvements over the baseline imputer. 
We note that the gain of MIRACLE is not reproducible via feature selection methods, which are still prone to the spurious correlations in the observed data, as discussed in Section \ref{subsect:problem}.

\vspace{-1mm}
\subsection{MIRACLE Convergence}
\begin{figure}[!t]
    \centering
    \vspace{-0.3cm}
    \includegraphics[width=\linewidth]{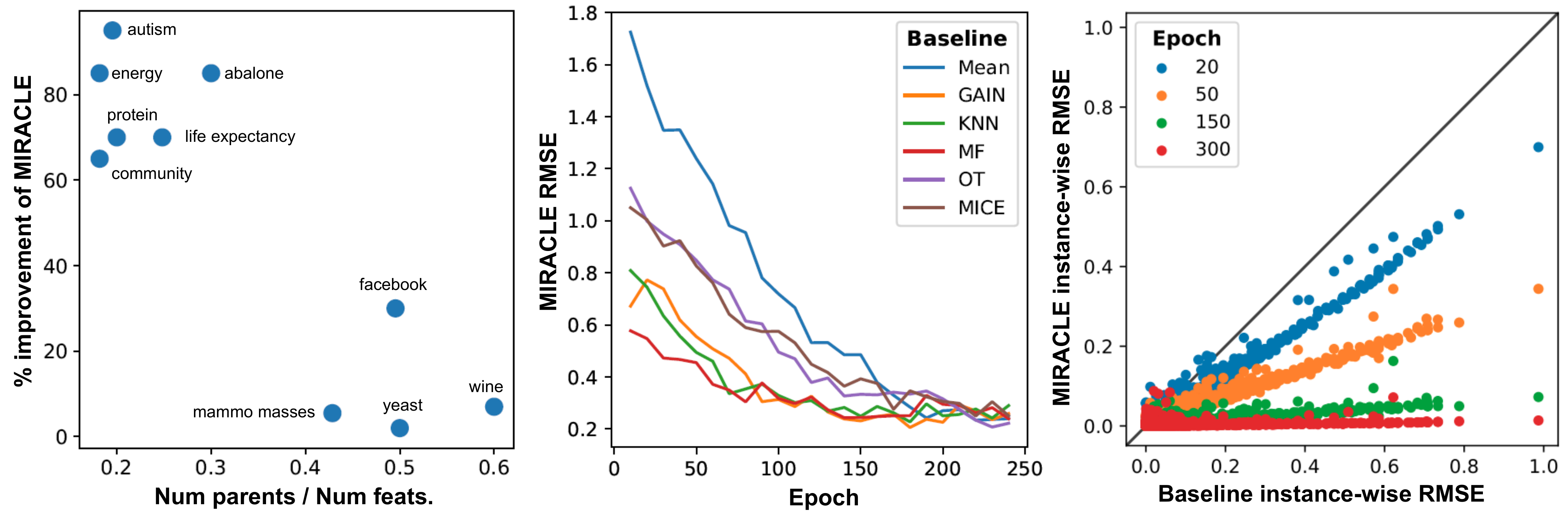}
    \vspace{-0.35cm}
    \caption{\textbf{(Left) Analysis of MIRACLE w.r.t. causal parents on real data.} MIRACLE has the most gain when we have identified a sparse set of causal parents.  
    When many features are identified as causal parents, imputation performance degrades.
    \textbf{(Mid) Convergence of MIRACLE across various baseline imputers.}
    On the abalone dataset, we show that MIRACLE converges to consistent RMSE regardless of baseline imputation.
    \textbf{(Right) Sample-wise RMSE for MIRACLE across various epochs.} MIRACLE is applied to refine MissForest imputations, demonstrating that error is reduced in a sample-wise basis.  Note: anything below the diagonal, is an improvement over the baseline imputations.
    }
    \label{fig:improvements}
    \vspace{-5mm}
\end{figure}

\vspace{-1mm}
In this subsection, we investigate two dimensions of MIRACLE refinement: (1) baseline imputation quality and (2) sample or instance-wise refinement. 
Regarding baseline imputation quality, we are interested in understanding the impact of MIRACLE refinement on various baseline imputers that may have disparate performances.  In the middle plot of Figure~\ref{fig:improvements}, we show MIRACLE applied to various baseline imputers on the Abalone dataset. Similar plots for other datasets can be found in Appendix~\ref{appx:additional_real_dataset}.  We observe that even though mean imputation starts off with the worst error, after refinement by MIRACLE, we see that all methods converge to similar RMSEs.
For the second experiment, we investigate the sample-wise improvement of MIRACLE on the abalone dataset using MissForest as a baseline imputer.  On the right-most plot of Figure~\ref{fig:improvements}, we observe that a vast majority of the samples are improved by MIRACLE. Note that every point below the diagonal is considered an improvement on an instance over the baseline imputation method. We can see MIRACLE improves the imputation almost universally except for the instances with small errors in the baseline imputation; on these instances, MIRACLE does not inflate their errors by a large margin. Furthermore, we observe that MIRACLE iteratively improves imputation as training progresses (over each epoch) by the observation that the slope of each line decreases with each epoch.


\section{Discussion} \label{sec:discussion}

In conclusion, motivated by the minimax optimization problem \eqref{robust_pop} arising from interventions on missingness indicators in the $m$-graph that encode the conditional independencies in the data distribution, we proposed MIRACLE, an iterative framework to refine any baseline missing data imputation to use the causal parents embodied in the estimated $m$-graphs. MIRACLE learns the causal $m$-graph as an adjacency matrix embedded in its input layers.  We proposed a two-part regularizer based on the causal graph and a moment regularizer based on the missing variable indicators.  We demonstrated that MIRACLE significantly improved the imputations of six baseline imputation methods over a variety of synthetic and real datasets. MIRACLE never hurts performance in the worst-case, and we envision MIRACLE becoming a de facto standard in refining missing data imputation.

There are several limitations we would like to identify as paths for future work. First,
any violation of the assumptions in Section~\ref{sect:back} may adversely impact the performance of 
MIRACLE in practice.
Second, causal discovery under missing data is an ongoing research area, and therefore MIRACLE may be discovering DAGs with bias introduced from the baseline methods.  However, in experiments, MIRACLE still performs well even if it starts with mean imputation. We expect MIRACLE to improve as causal discovery methods under missingness improve.
Third, in its current form, MIRACLE is not extensible to scenarios where causality may not be applicable, such as computer vision.
Fourth, because of the causal discovery regularizer and network architecture, MIRACLE may have difficulty scaling to very high dimensional data.  
Lastly, a more general and detailed discussion is needed between our work and the merits of causality and robustness.

\section*{Acknowledgements}
This work was supported by \textit{GlaxoSmithKline} (GSK), the \textit{National Science Foundation} (NSF) under grant number 1722516, the \textit{Office of Naval Research} (ONR), and \textit{The Alan Turning Institute} (ATI).
We thank all reviewers for their invaluable comments and suggestions.

\bibliographystyle{plain}
\bibliography{main_neurips}

\clearpage
\newpage

\section*{Checklist}

\begin{enumerate}

\item For all authors...
\begin{enumerate}
  \item Do the main claims made in the abstract and introduction accurately reflect the paper's contributions and scope?
    \answerYes{}
  \item Did you describe the limitations of your work?
    \answerYes{See the Discussion in Section~\ref{sec:discussion}.}
  \item Did you discuss any potential negative societal impacts of your work?
    \answerNA{We do not foresee any negative societal impacts of our work.}
  \item Have you read the ethics review guidelines and ensured that your paper conforms to them?
    \answerYes{}
\end{enumerate}

\item If you are including theoretical results...
\begin{enumerate}
  \item Did you state the full set of assumptions of all theoretical results?
    \answerNA{}
	\item Did you include complete proofs of all theoretical results?
    \answerNA{}
\end{enumerate}

\item If you ran experiments...
\begin{enumerate}
  \item Did you include the code, data, and instructions needed to reproduce the main experimental results (either in the supplemental material or as a URL)?
    \answerYes{The code is available at \url{https://anonymous.4open.science/r/MIRACLE-7702/}.  The data is all publicly available at \cite{uci}.  The instructions to reproduce our work can be found in the experimental Section~\ref{section:experiments} and Appendix~\ref{appx:synth_DGP}.}
  \item Did you specify all the training details (e.g., data splits, hyperparameters, how they were chosen)?
    \answerYes{See Section~\ref{section:experiments}.}
	\item Did you report error bars (e.g., with respect to the random seed after running experiments multiple times)?
    \answerYes{See Section~\ref{section:experiments}.}
	\item Did you include the total amount of compute and the type of resources used (e.g., type of GPUs, internal cluster, or cloud provider)?
    \answerYes{See Appendix~\ref{appx:training_details}.}
\end{enumerate}

\item If you are using existing assets (e.g., code, data, models) or curating/releasing new assets...
\begin{enumerate}
  \item If your work uses existing assets, did you cite the creators?
    \answerYes{}
  \item Did you mention the license of the assets?
    \answerNA{}
  \item Did you include any new assets either in the supplemental material or as a URL?
    \answerNA{}
  \item Did you discuss whether and how consent was obtained from people whose data you're using/curating?
    \answerNA{}
  \item Did you discuss whether the data you are using/curating contains personally identifiable information or offensive content?
    \answerNA{}
\end{enumerate}

\item If you used crowdsourcing or conducted research with human subjects...
\begin{enumerate}
  \item Did you include the full text of instructions given to participants and screenshots, if applicable?
    \answerNA{}
  \item Did you describe any potential participant risks, with links to Institutional Review Board (IRB) approvals, if applicable?
    \answerNA{}
  \item Did you include the estimated hourly wage paid to participants and the total amount spent on participant compensation?
    \answerNA{}
\end{enumerate}

\end{enumerate}


\onecolumn
\appendix

\textbf{\Large Appendix}
\\\\

This appendix is outlined as follows:
\begin{itemize}
    \item Section~\ref{appx:synth_DGP} details our synthetic data generating process and how we generated missingness.
    \item Section \ref{appdx:additionalsynth} contains additional experiments on synthetic data testing for imputation performance, testing for prediction performance of a downstream machine learning algorithm on imputed data, and testing for congeniality. In each case we consider MCAR, MAR and MNAR missingness patterns.  We used the same experimental setup in \cite{GAIN} for testing prediction performance of a downstream machine learning algorithm and for congeniality.  
    \item Section~\ref{appx:additional_real_dataset} contains additional experiments on real datasets testing for prediction performance of a downstream machine learning algorithm on imputed data, and testing for congeniality over all types of missingness.
    \item Section~\ref{appx:training_details} provides an overview of our model and training details.
    \item Section~\ref{appdx:complexity} provides a computational complexity analysis of MIRACLE.
    \item Section~\ref{appx:ablation} provides an ablation study for the MIRACLE loss function.
    \item Section~\ref{apx:missingness} provides experiments regarding missingness location.
    
\end{itemize}

\newpage

\section{Supplementary experimental details}\label{appx:synth_DGP}

\subsection{Synthetic data generation}
In each synthetic experiment, we generated a $p$-dimensional random graph $G$ from a Erd\"os–R\'enyi random graph model with $p$ edges on average. Given $G$, we assigned uniformly random edge weights to obtain a weighted adjacency matrix $W\in\mathbb R^{p\times p}$. Given $W$, we sampled $X = W X  + E$ repeatedly from a Gaussian noise model for $E\in\mathbb R^p$ (each dimension sampled independently) to generate independent observations from this system.

\subsection{Generating missingness.}\label{appx:missingness}

The following explains how we constructed synthetic datasets that satisfy MCAR, MAR and MNAR patterns of missingness. We apply a modification to the missingness generation from \cite{GAIN}.
\begin{itemize}
    \item \textbf{MCAR}. Missing completely at random was introduced by randomly removing $30\%$ of the observations in each feature.
    \item \textbf{MAR}. 
    We sequentially define the probability that the $i$-th component of the $n$-th sample is observed conditional on the missingness and values (if observed) of the previous $i - 1$ components to be,
    \begin{align}
        P^m(i) = \frac{p^m(i) \cdot N \cdot \exp(\sum_{j<i}w_jm_j(n)x_j(n)+b_j (1 - m_j(n)) )}{\sum_{l=1}^N\exp(\sum_{j<i}w_jm_j(l)x_j(l)+b_j (1 - m_j(l)) )}
    \end{align}
    where $p^m(i)$ corresponds to the average missing rate of the $i$-th feature, and $w_j$, $b_j$ are sampled from $\mathcal U(0, 1)$ (but are only sampled once for the entire dataset). 
    \item \textbf{MNAR.} 
    Missing not at random was introduced by defining the probability of the $i$-th component of the $n$-th sample to be observed by,
    \begin{align}
    P^m(i) = \frac{p^m(i) \cdot N \cdot \exp(-w_ix_i(n))}{\sum_{l=1}^N\exp(-w_ix_i(l))}
    \end{align}
    with the same notation as above. Here, the missingness of a data point is directly dependent on its value (with dependence determined by the weight $w_i$, sampled from $\mathcal U(0,1)$).
\end{itemize}

\section {Additional synthetic results} \label{appdx:additionalsynth}

In this section, we provide supplementary results for synthetic experiments.  Note that the error bars are large for some of the plots with predictive error and congeniality.  This is because the y-axis of these plots are min-max normalized between 0 and 1, so the high variance (large error bars) shows that the improvement by MIRACLE may be minimal for the mentioned datasets.  Additionally, this could be caused by the fact that the missing features aren't predictive of a target variable, i.e., better imputation does not necessarily lead to any performance gain for the predicting the target variable.



\subsection{MCAR Results}
\begin{figure}[!htbp]
    \centering
    \subfloat[Performance in terms of RMSE.]{\includegraphics[width=0.45\linewidth]{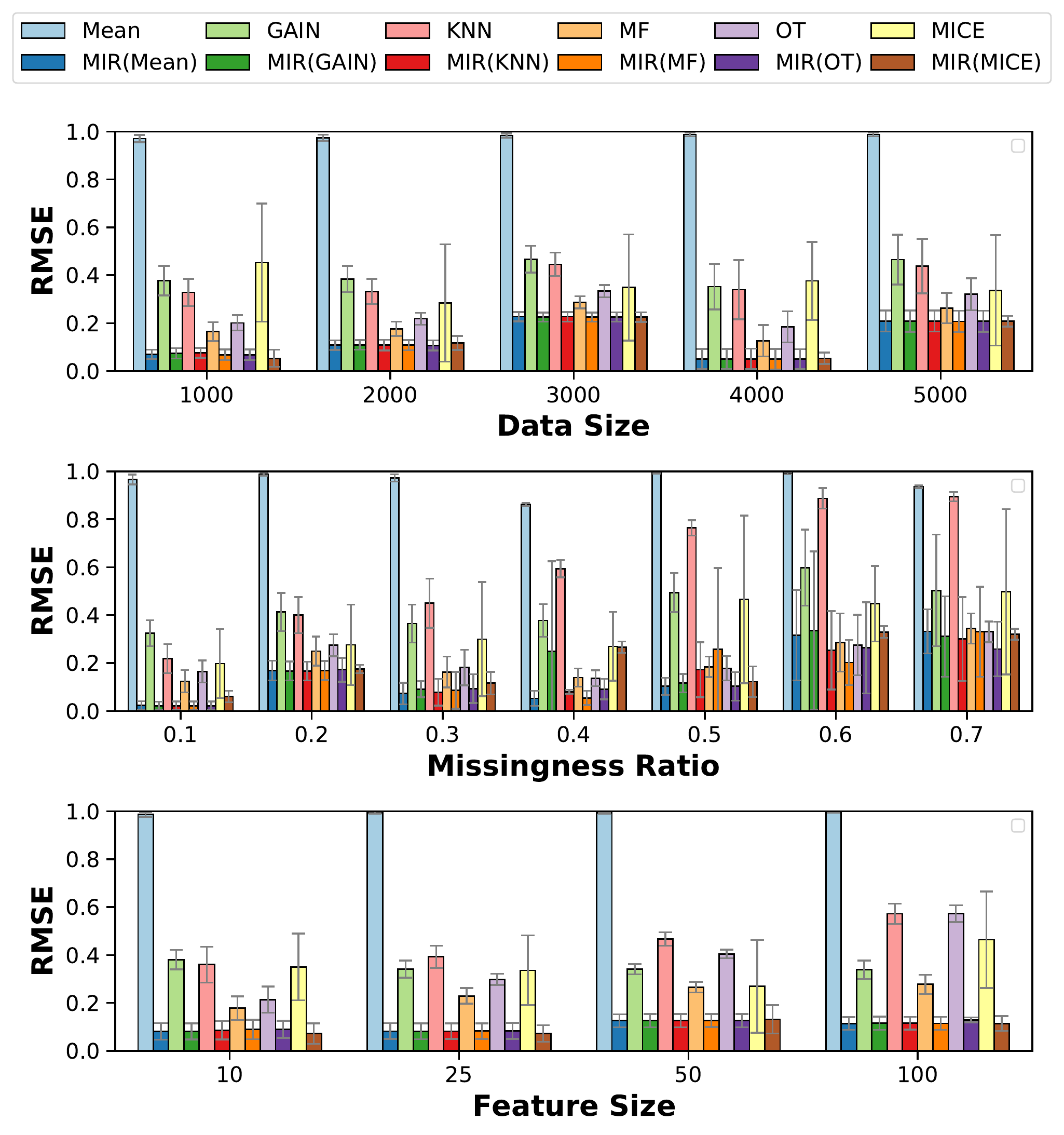} }
    \qquad
    \subfloat[Performance in terms of prediction RMSE.]{\includegraphics[width=0.45\linewidth]{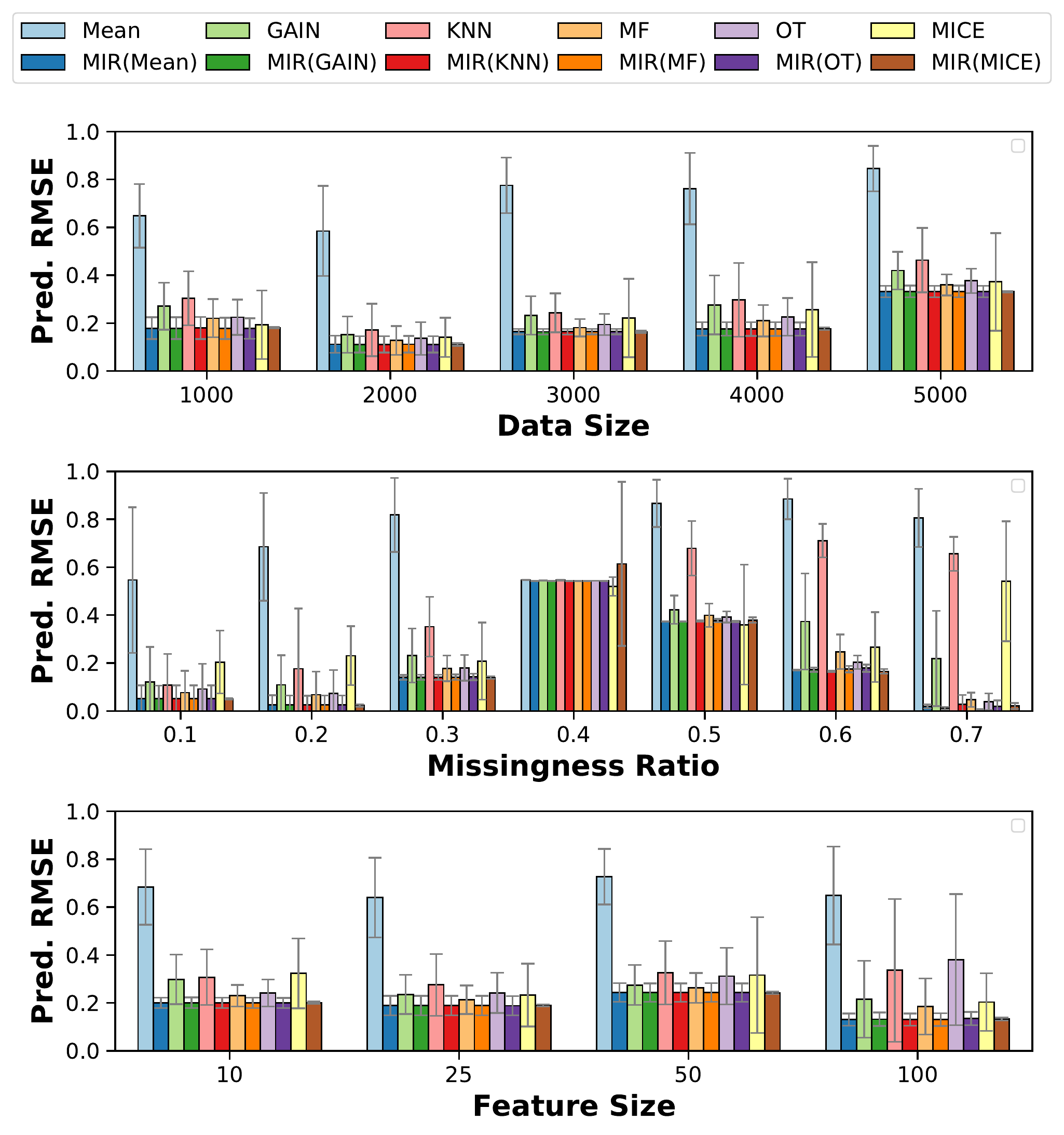}}
    \qquad
    \subfloat[MCAR congeniality (in terms of RMSE).]{\includegraphics[width=0.45\linewidth]{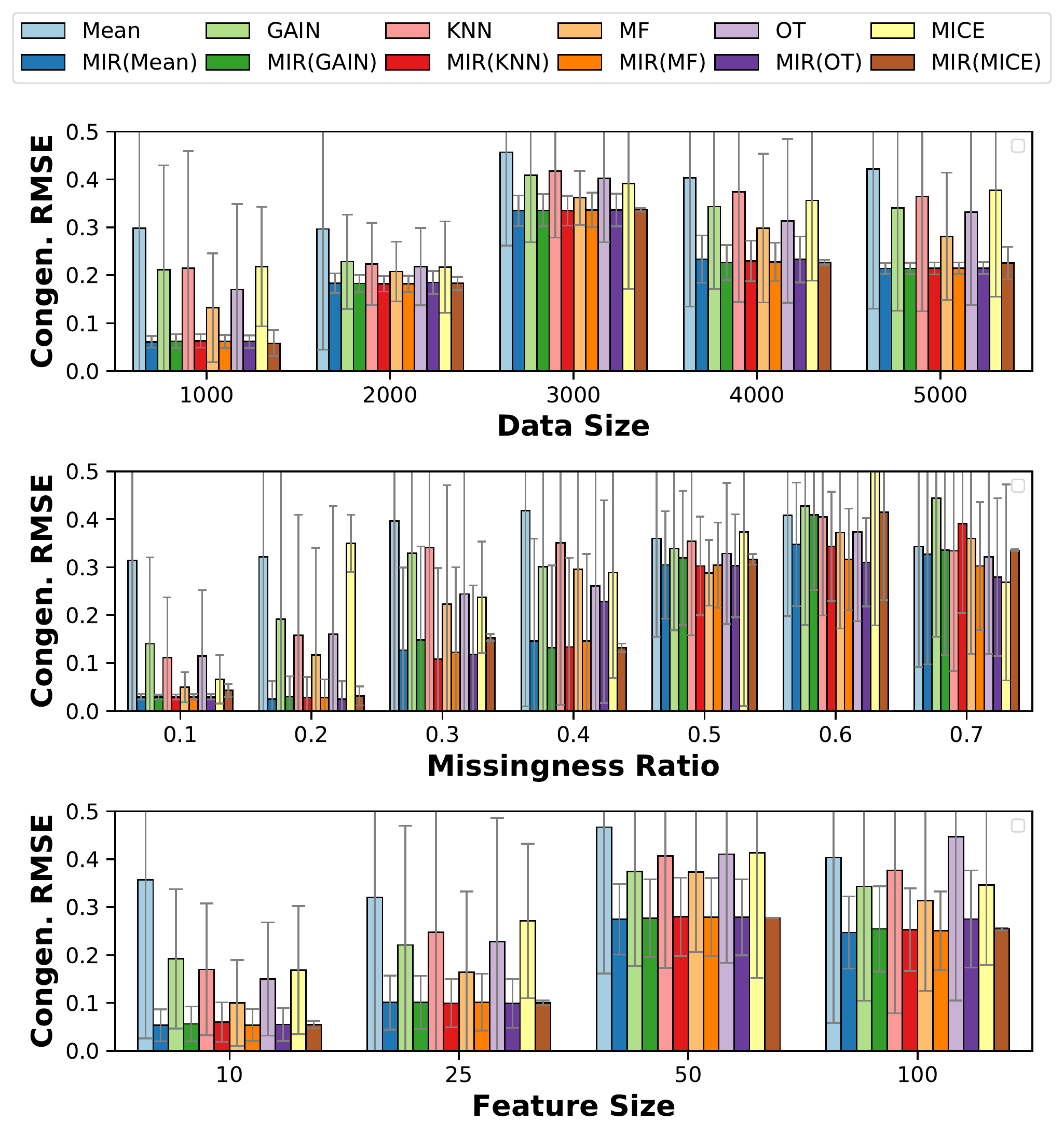}}
    \caption{Experiments on MCAR synthetic data as a function of dataset sizes \textbf{(top)}, missingness rates \textbf{(middle)}, and feature sizes \textbf{(bottom)} of each subfigure: \textbf{(a)} RMSE, \textbf{(b)} machine learning predictive error  of a random variable, and \textbf{(c) }congeniality. }
    \label{fig:MCAR_app}
\end{figure}

Using our synthetic experimental setup used in the main paper, we show the performance of MIRACLE in terms of RMSE, predictive error, and congeniality in Figure~\ref{fig:MCAR_app} for each of our baseline methods with MCAR.

\newpage
\subsection{MAR Results}

\begin{figure}[!htbp]
    \centering
    \subfloat[Performance in terms of RMSE.]{\includegraphics[width=0.45\linewidth]{figures/synth_err_MAR.pdf} }%
    \qquad
    \subfloat[Performance in terms of prediction RMSE.]{\includegraphics[width=0.45\linewidth]{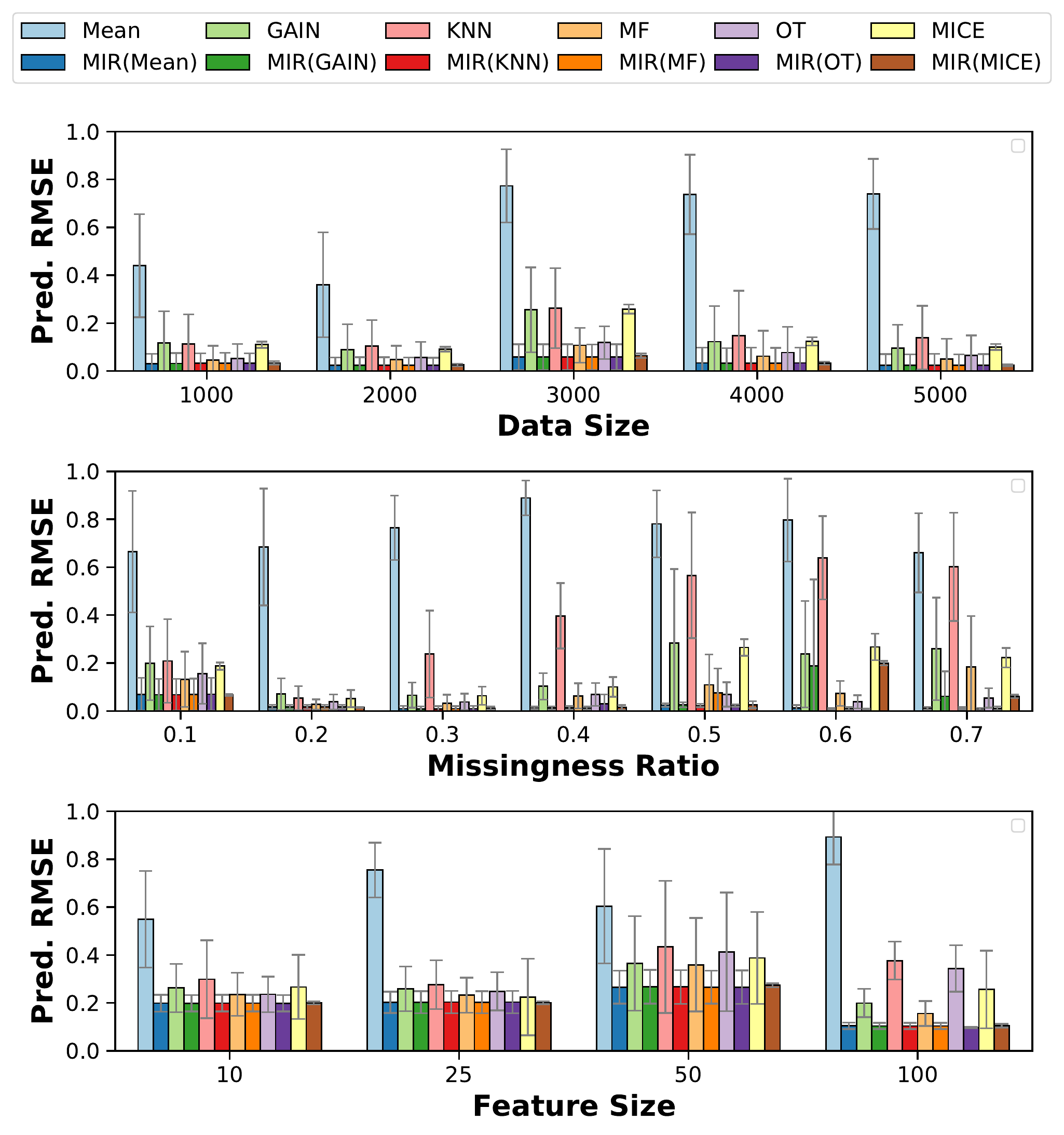}}
    \qquad
    \subfloat[Congeniality (in terms of RMSE).]{\includegraphics[width=0.45\linewidth]{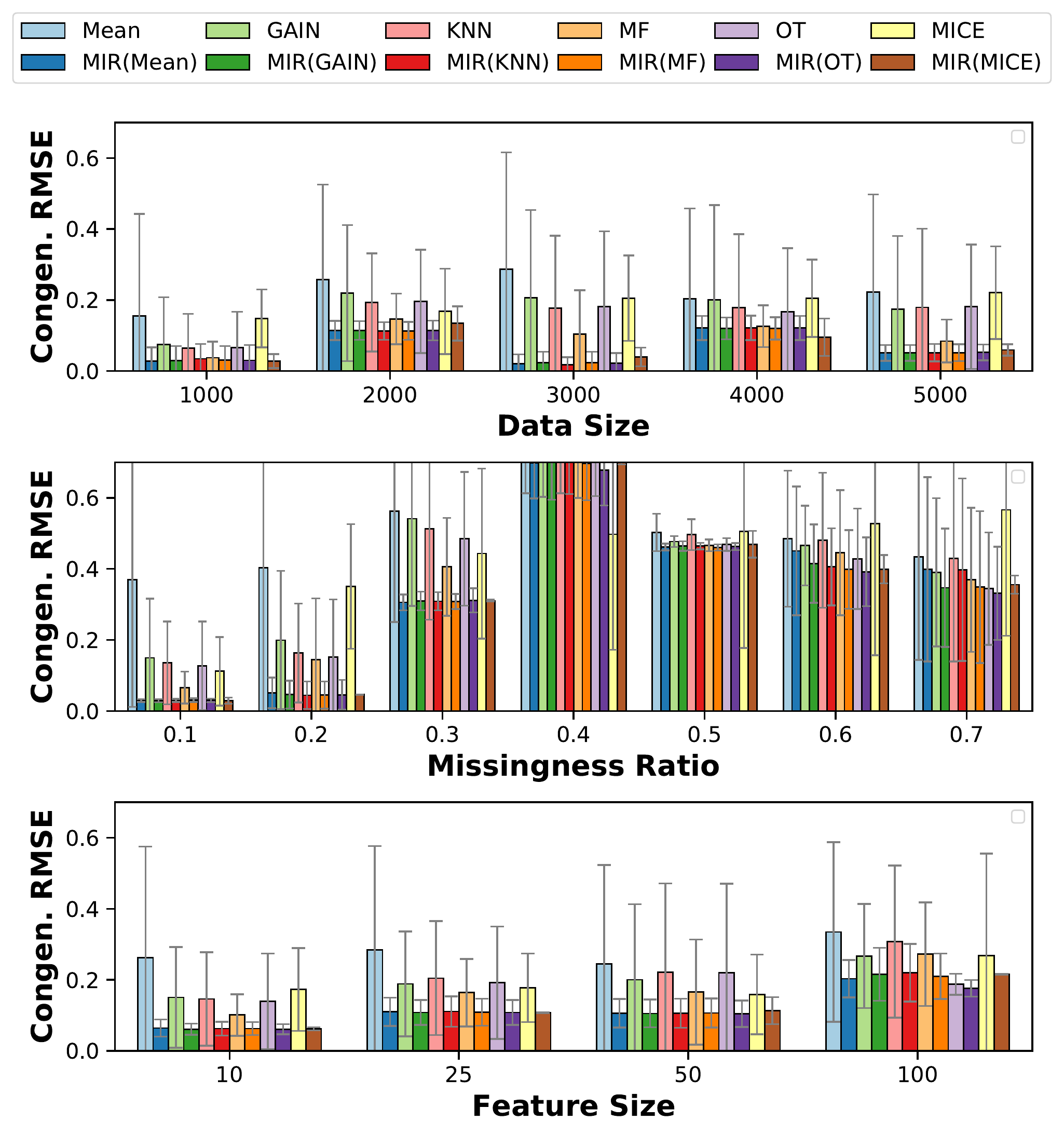}}
    \caption{Experiments on MAR synthetic data as a function of dataset sizes \textbf{(top)}, missingness rates \textbf{(middle)}, and feature sizes \textbf{(bottom)} of each subfigure: \textbf{(a)} RMSE, \textbf{(b)} machine learning predictive error  of a random variable, and \textbf{(c) }congeniality.}
    \label{fig:MAR_app}
\end{figure}

Using our synthetic experimental setup used in the main paper, we show the performance of MIRACLE in terms of RMSE, predictive error, and congeniality in Figure~\ref{fig:MAR_app} for each of our baseline methods with MAR.

\subsection{MNAR Results}

\begin{figure}[!htbp]
    \centering
    \subfloat[Performance in terms of RMSE.]{\includegraphics[width=0.45\linewidth]{figures/synth_err_MNAR.pdf} }%
    \qquad
    \subfloat[Performance in terms of prediction RMSE.]{\includegraphics[width=0.45\linewidth]{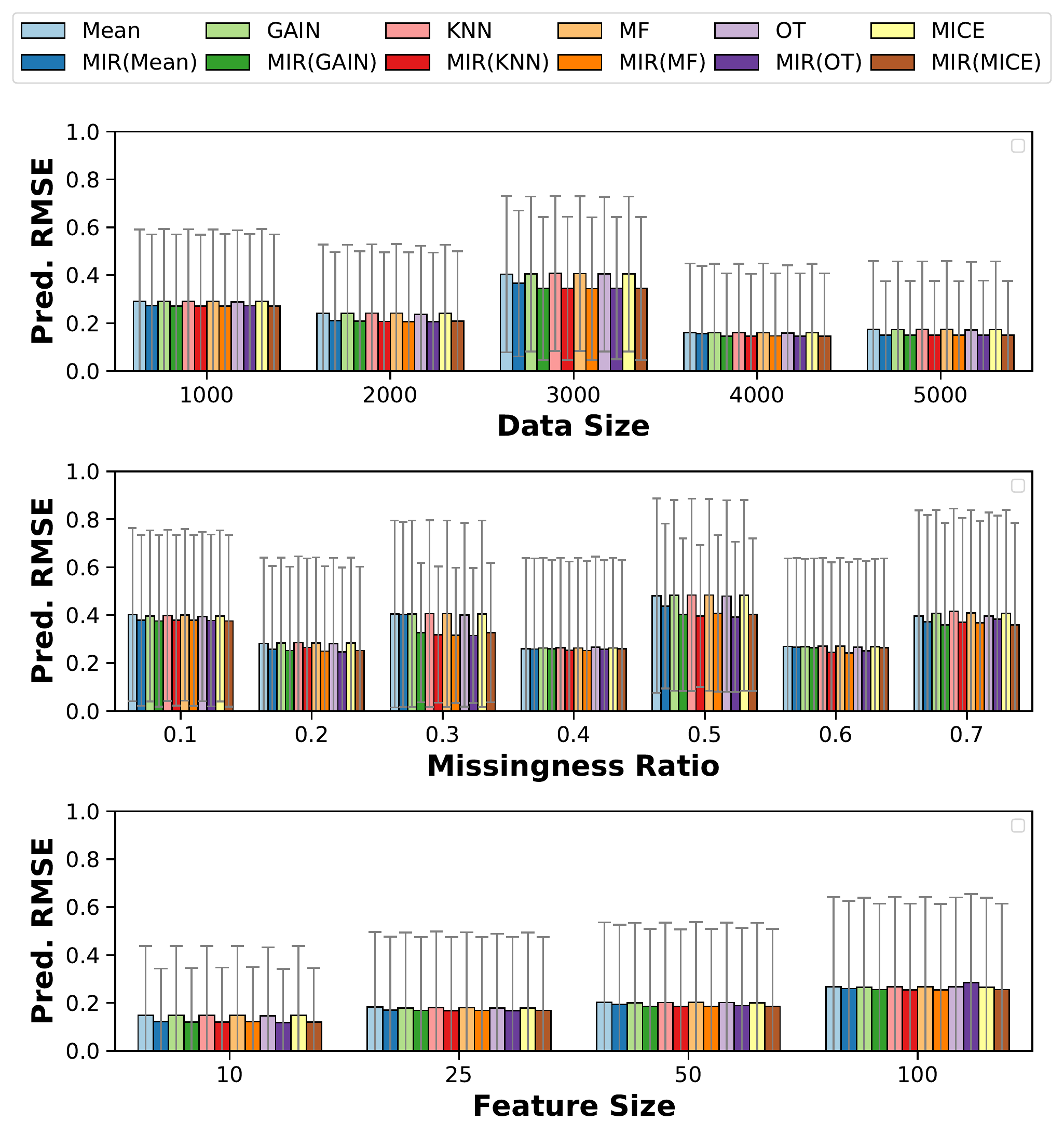}}
    \qquad
    \subfloat[Congeniality (in terms of RMSE).]{\includegraphics[width=0.45\linewidth]{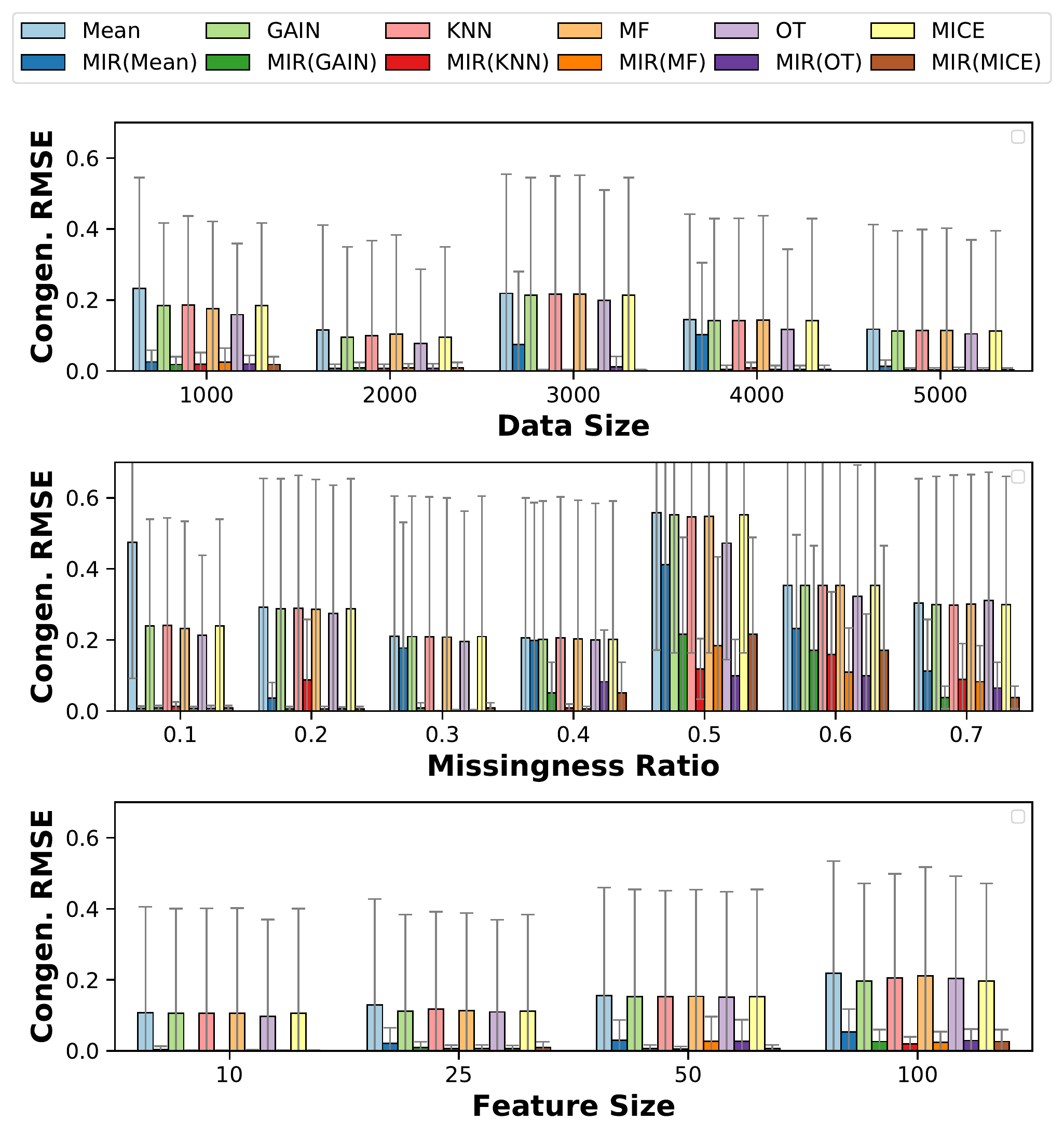}}
    \caption{Experiments on MNAR synthetic data as a function of dataset sizes \textbf{(top)}, missingness rates \textbf{(middle)}, and feature sizes \textbf{(bottom)} of each subfigure: \textbf{(a)} RMSE, \textbf{(b)} machine learning predictive error  of a random variable, and \textbf{(c) }congeniality. }
    \label{fig:MNAR_app}
\end{figure}

Using our synthetic experimental setup used in the main paper, we show the performance of MIRACLE in terms of RMSE, predictive error, and congeniality in Figure~\ref{fig:MNAR_app} for each of our baseline methods with MNAR.

\newpage
\section{Additional real datasets}\label{appx:additional_real_dataset}

\paragraph{Prediction error and congeniality.} We include additional plots for the real data experiments for prediction error and congeniality in Figures~\ref{fig:real_prederr} and \ref{fig:real_congen}, respectively.

\begin{figure}[!ht]
\centering
    \begin{subfigure}{0.3\linewidth}
        \includegraphics[width=\linewidth]{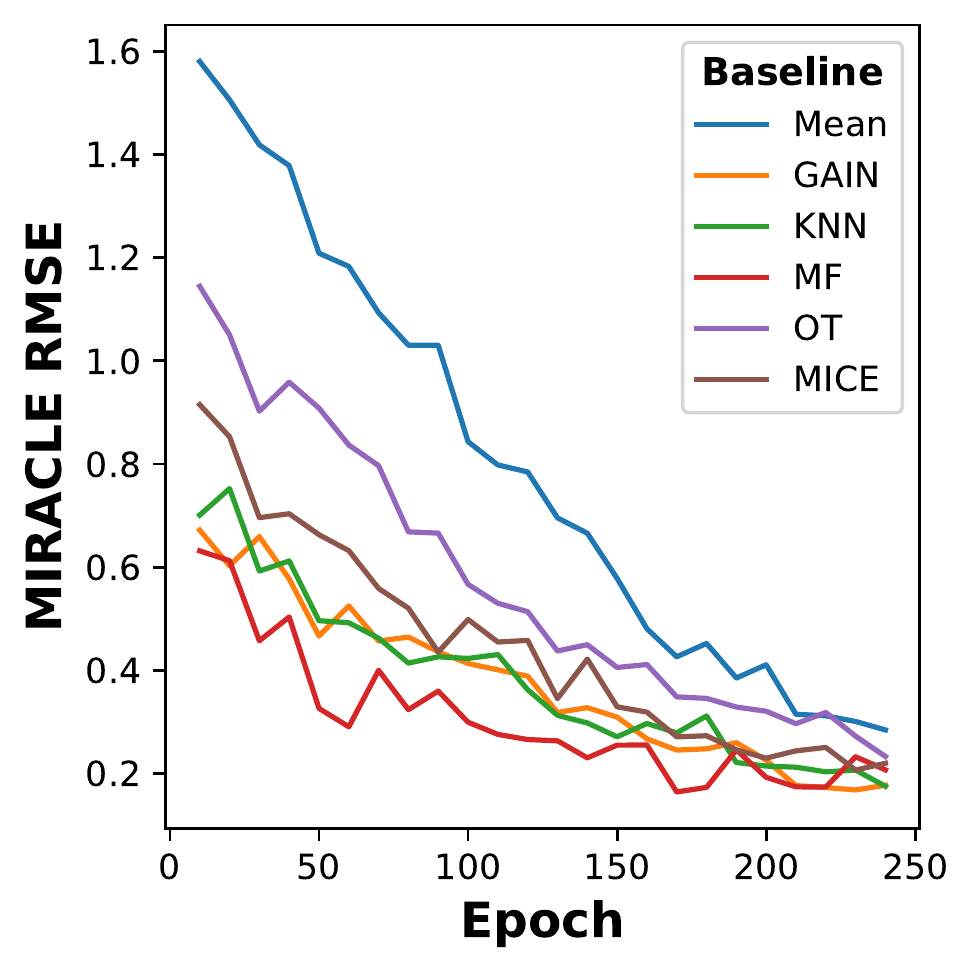}
    \caption{Autism}
    \end{subfigure}
\hfil
    \begin{subfigure}{0.3\linewidth}
        \includegraphics[width=\linewidth]{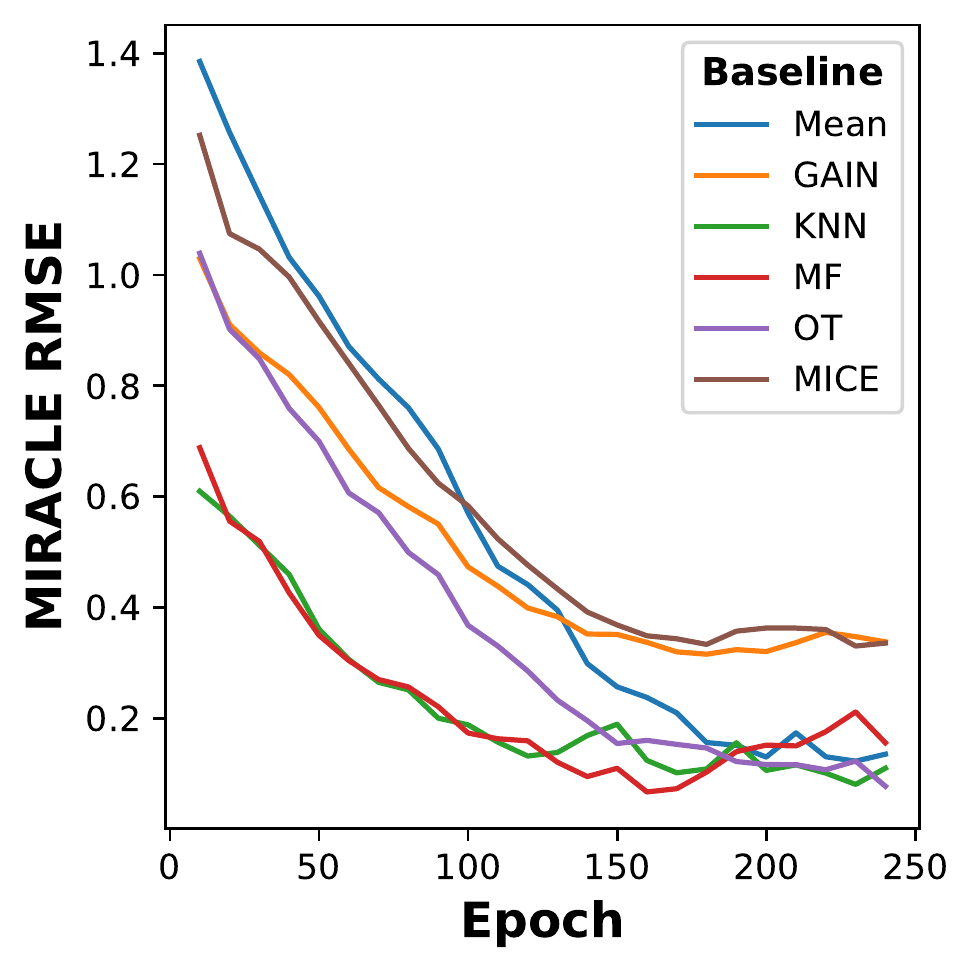}
    \caption{Energy}
    \end{subfigure}
    \hfil
    \begin{subfigure}{0.3\linewidth}
        \includegraphics[width=\linewidth]{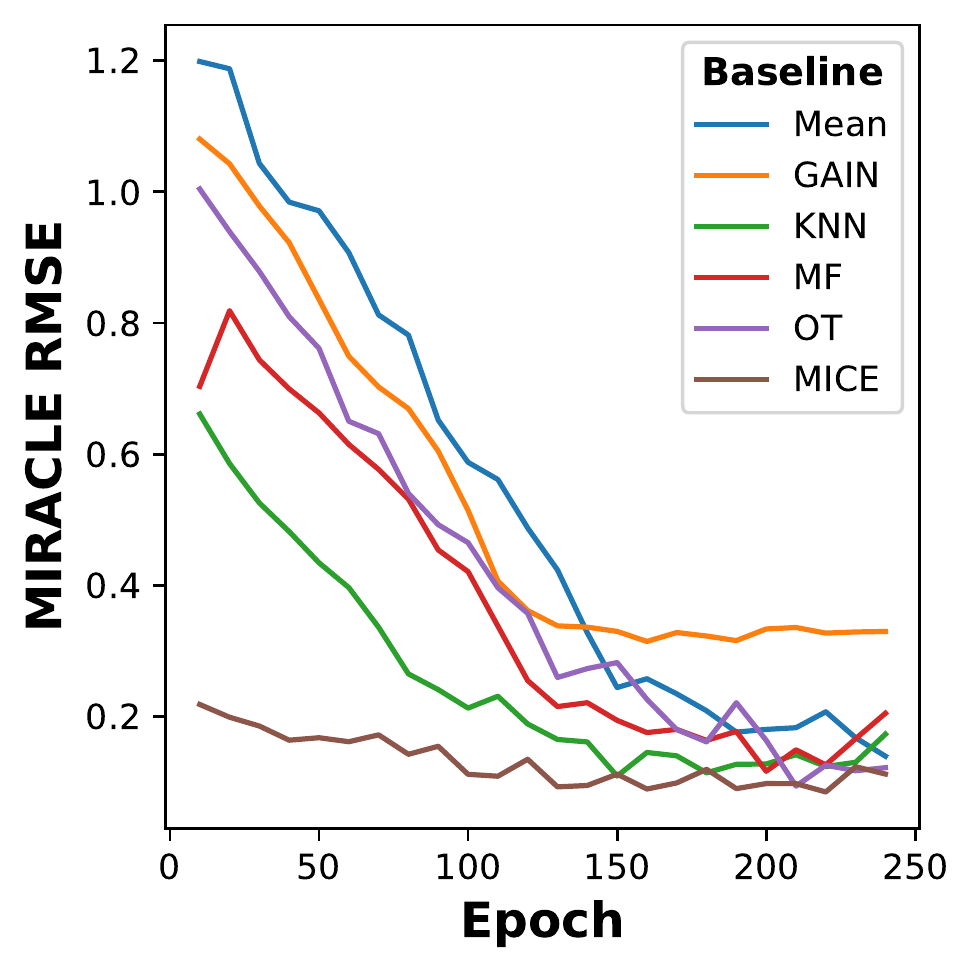}
    \caption{Protein}
    \end{subfigure}

    \begin{subfigure}{0.3\linewidth}
        \includegraphics[width=\linewidth]{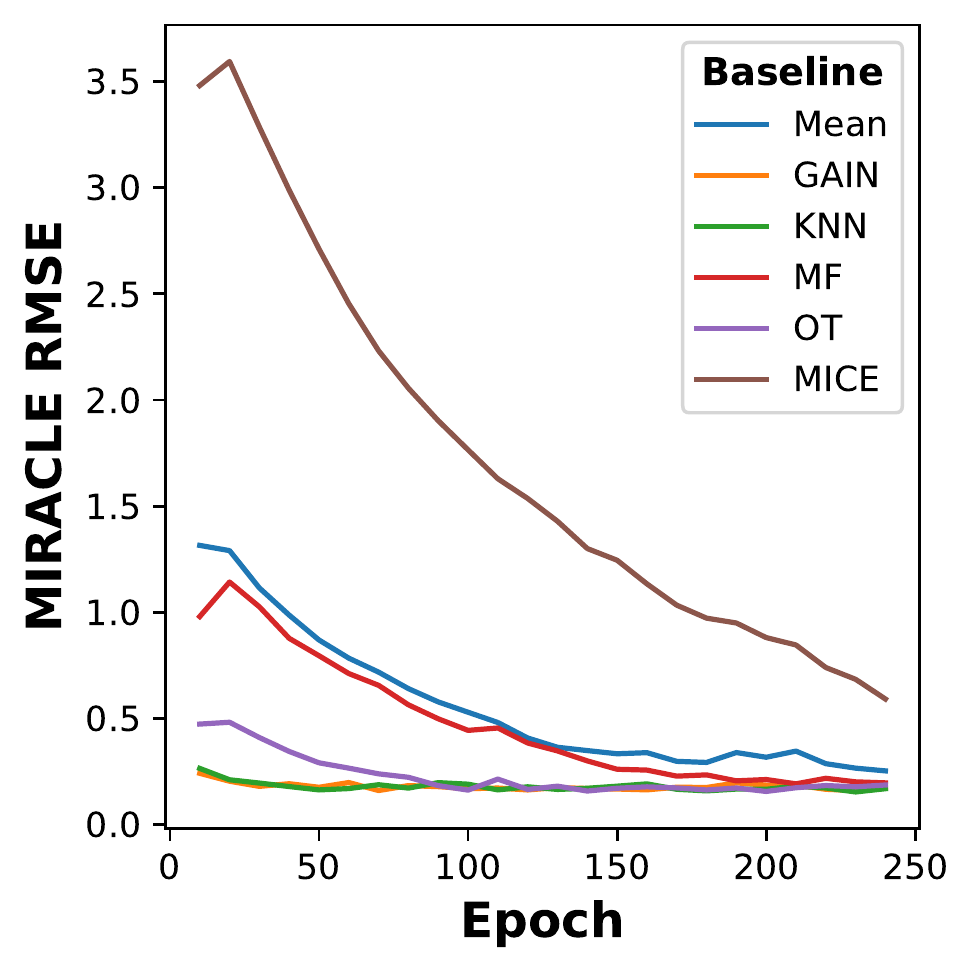}
    \caption{Life expectancy}
    \end{subfigure}
\hfil
    \begin{subfigure}{0.3\linewidth}
        \includegraphics[width=\linewidth]{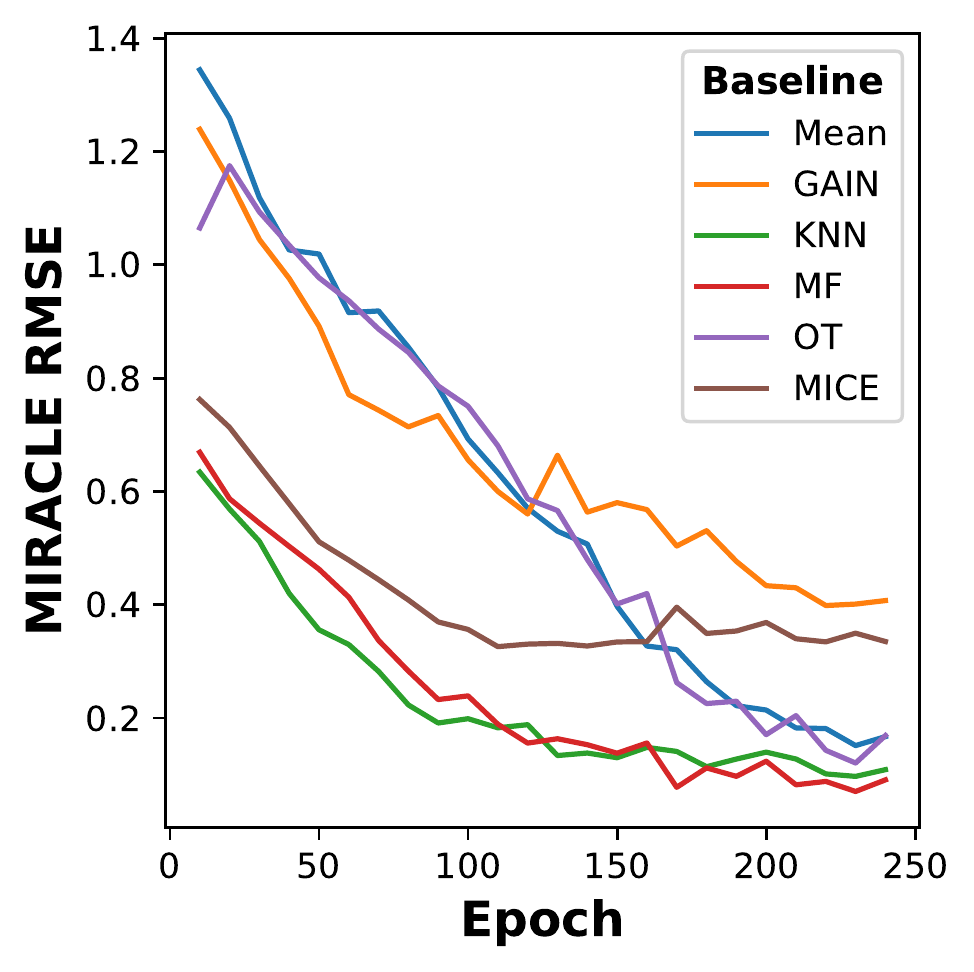}
    \caption{Community}
    \end{subfigure}
    \hfil
    \begin{subfigure}{0.3\linewidth}
        \includegraphics[width=\linewidth]{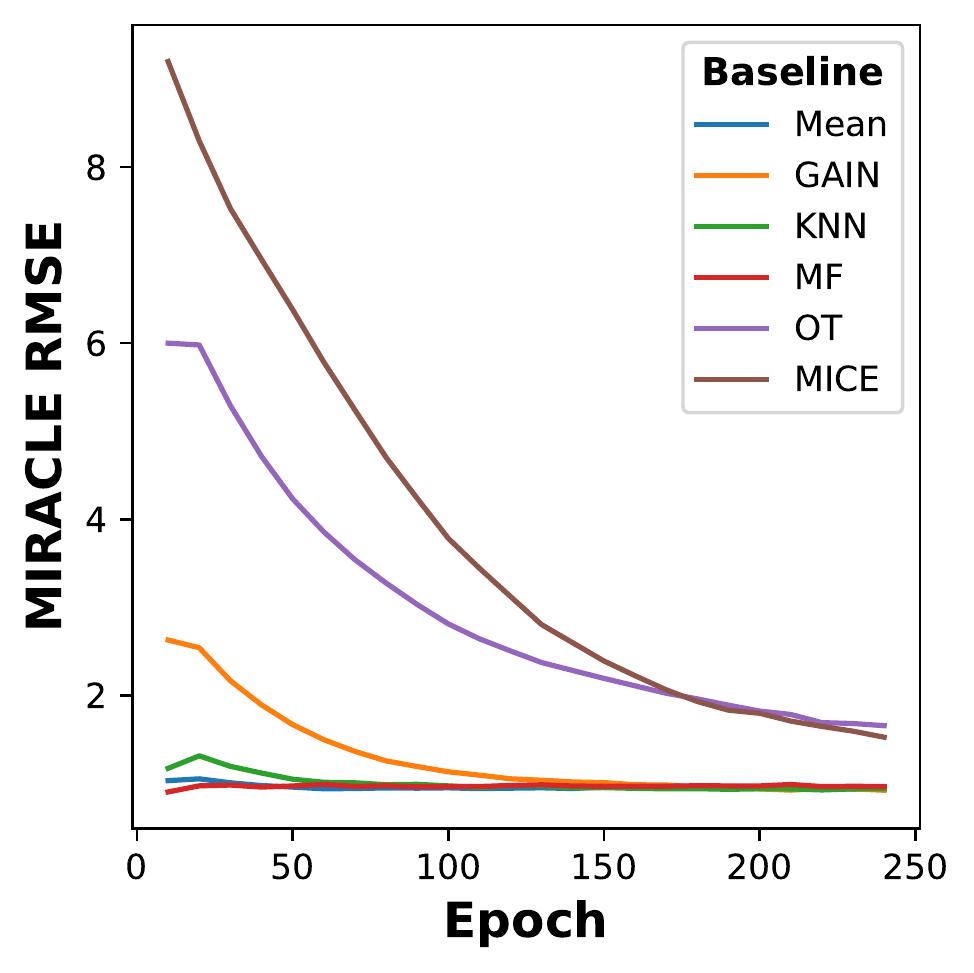}
    \caption{Yeast}
    \end{subfigure}

        \begin{subfigure}{0.3\linewidth}
        \includegraphics[width=\linewidth]{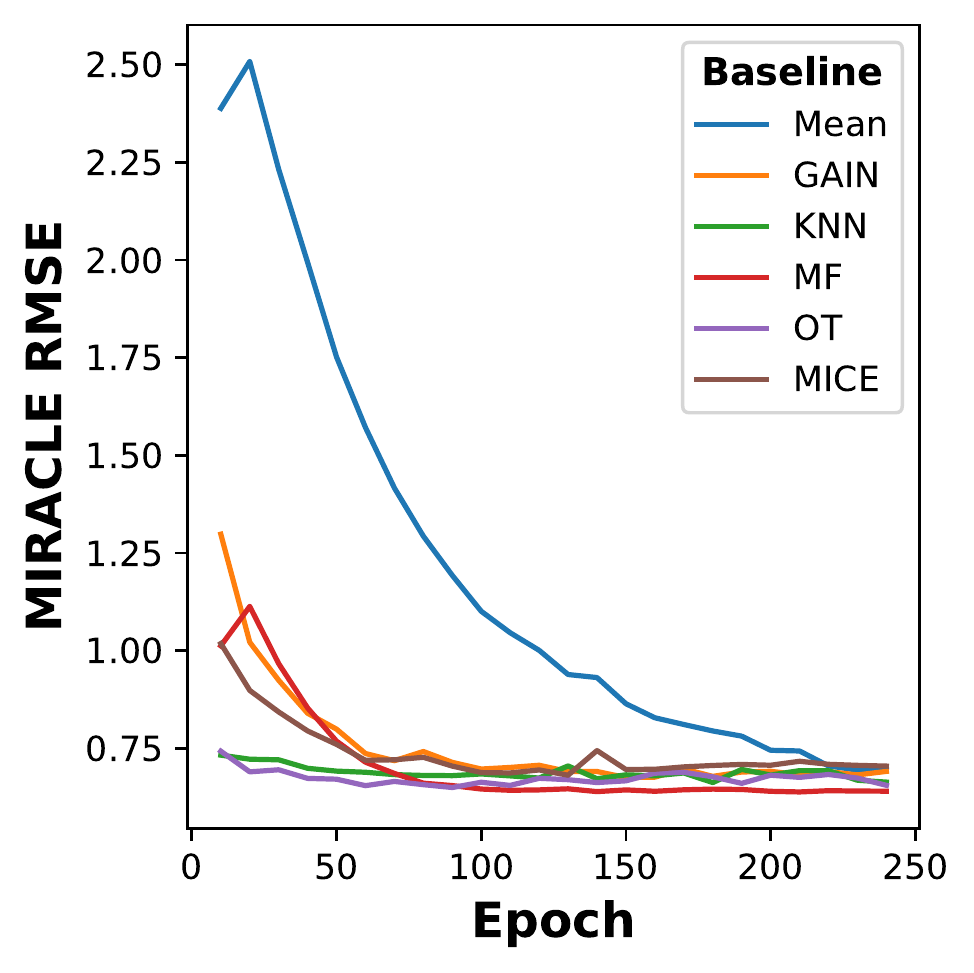}
    \caption{Mammo. masses}
    \end{subfigure}
\hfil
    \begin{subfigure}{0.3\linewidth}
        \includegraphics[width=\linewidth]{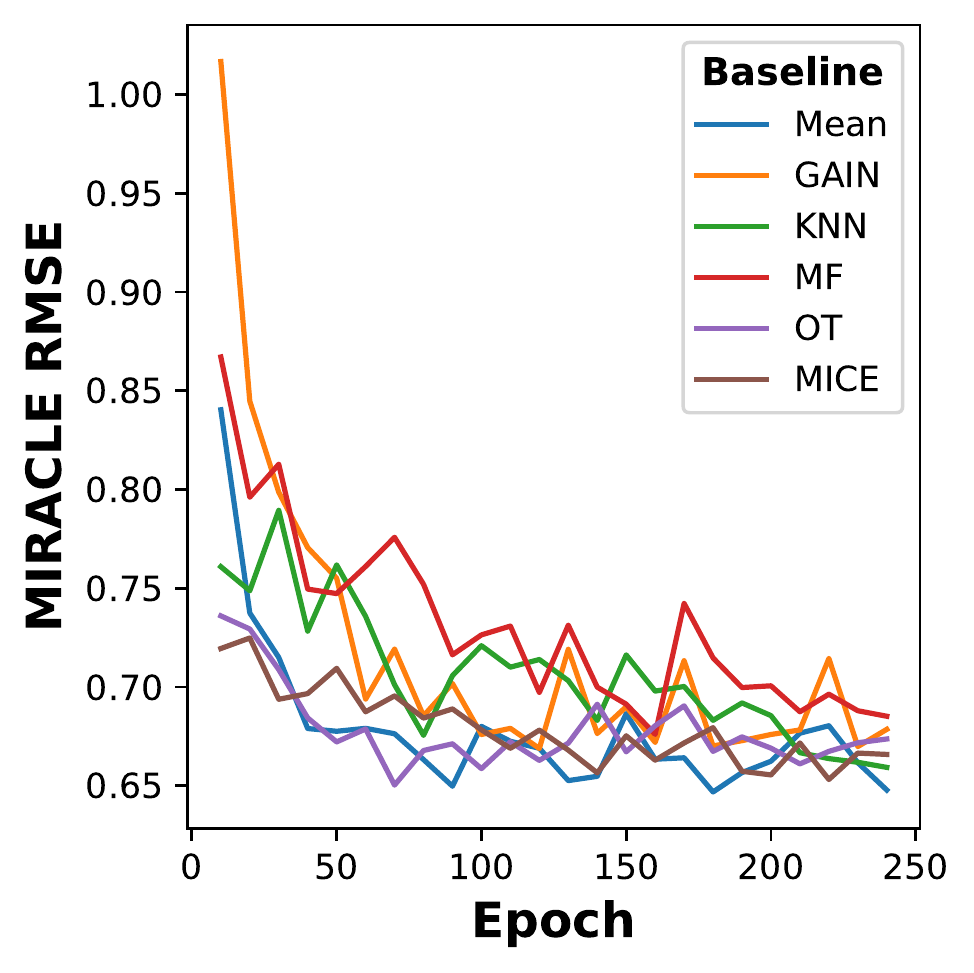}
    \caption{Wine quality}
    \end{subfigure}
    \hfil
    \begin{subfigure}{0.3\linewidth}
        \includegraphics[width=\linewidth]{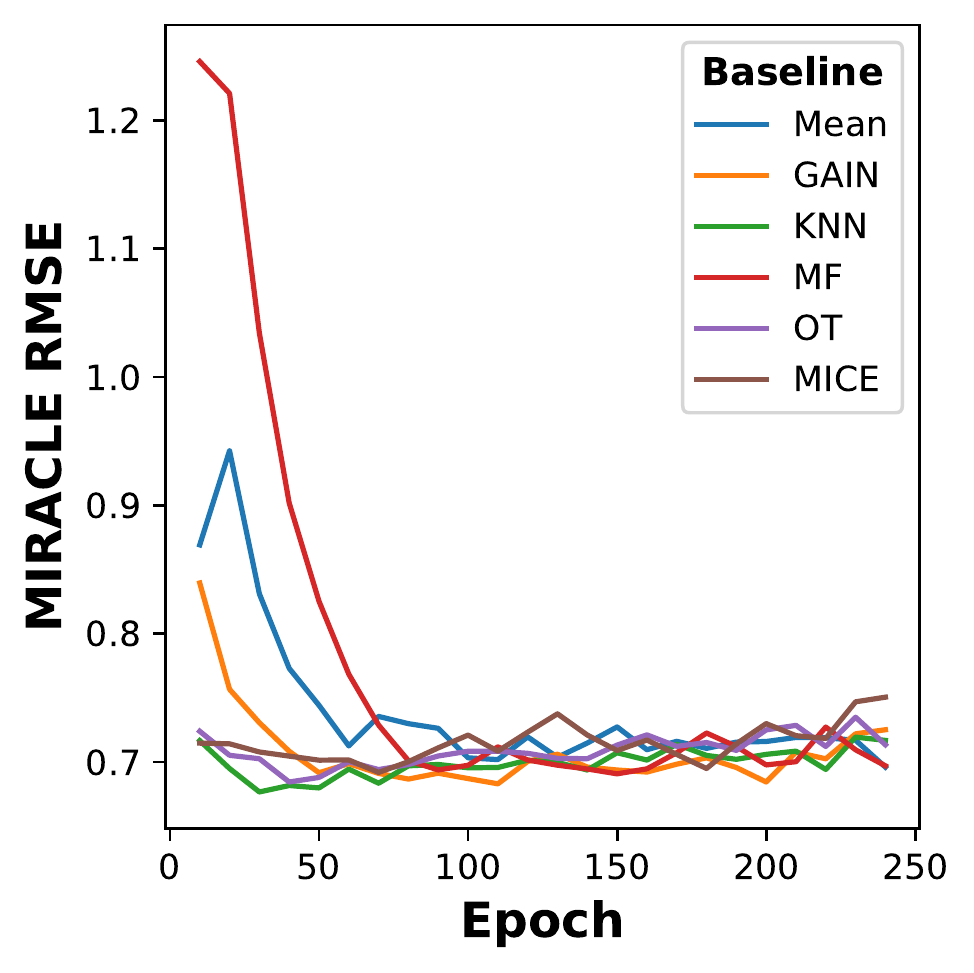}
    \caption{Facebook}
    \end{subfigure}
\caption{Convergence plots for real datasets.}
    \label{fig:convergenceplots}
    \end{figure}

\begin{figure*}[!htbp]
    \centering
    \includegraphics[width=1\linewidth]{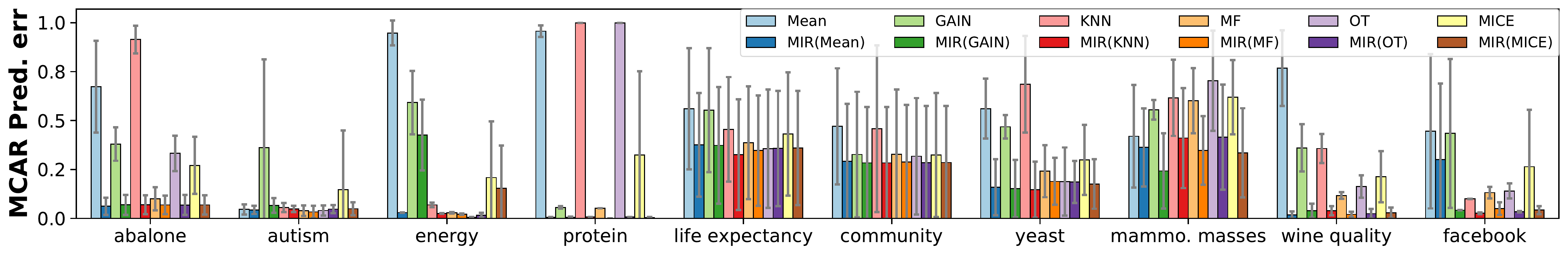} 
    \includegraphics[width=1\linewidth]{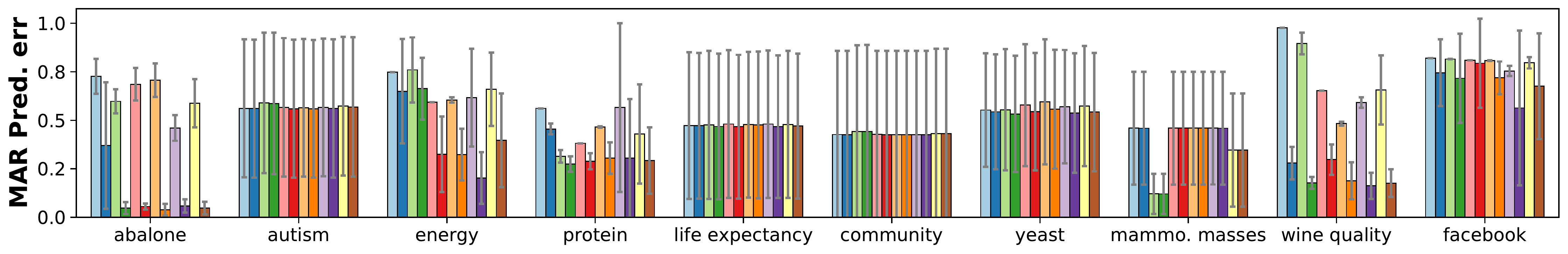} 
    \includegraphics[width=1\linewidth]{figures/real_prederr_MAR.pdf} 
    \caption{\textbf{MIRACLE on real datasets in terms of predictive error}.  MIRACLE improves over all baselines across all types of missingness: MCAR \textbf{(top)}, MAR \textbf{(middle)}, and MNAR \textbf{(bottom)}. }
    \label{fig:real_prederr}
\end{figure*}

\begin{figure*}[!htbp]
    \centering
        \includegraphics[width=1\linewidth]{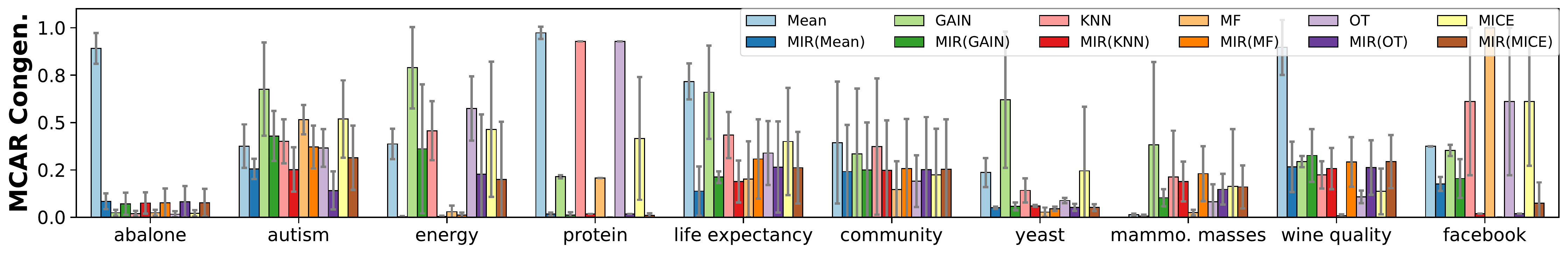} 
    \includegraphics[width=1\linewidth]{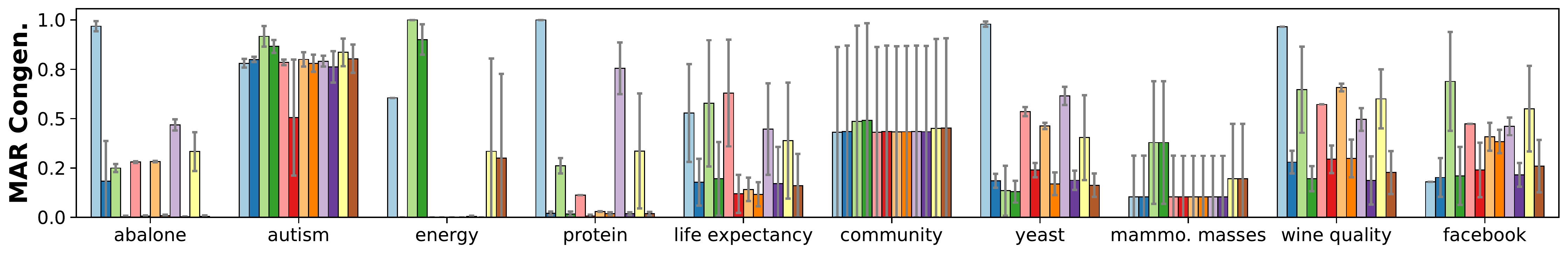} 
    \includegraphics[width=1\linewidth]{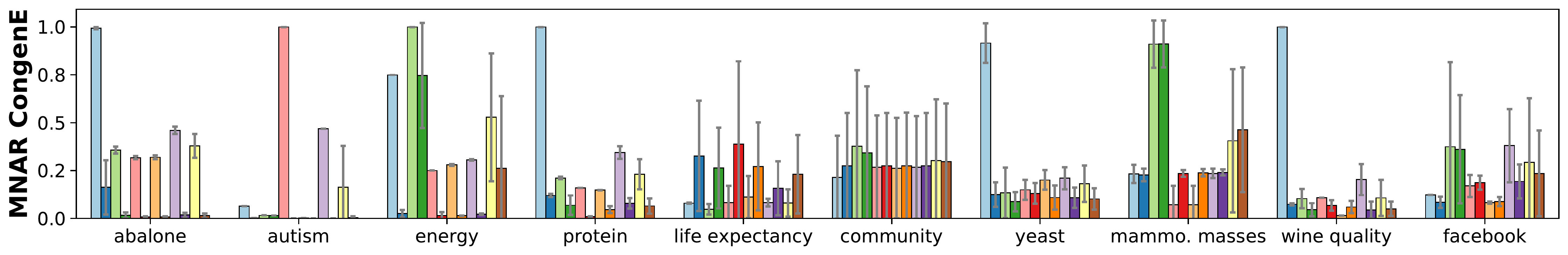} 
    \caption{\textbf{MIRACLE on real datasets in terms of congeniality}.  MIRACLE improves over all baselines across all types of missingness: MCAR \textbf{(top)}, MAR \textbf{(middle)}, and MNAR \textbf{(bottom)}. }
    \label{fig:real_congen}
\end{figure*}

\paragraph{Additional convergence plots.} We include additional convergence plots on real datasets in Figure~\ref{fig:convergenceplots}. We use the same experimental setup used in Figure~\ref{fig:improvements} in Section~\ref{section:experiments}.  We observe that MIRACLE is able to converge regardless of baseline imputation used.

\section{Model and training details}\label{appx:training_details}

We used the following network architecture for MIRACLE.  Our proposed architecture consists of $d$ sub-networks with shared hidden layers, as shown in Figure~\ref{fig:schematic}. 
Each network is constructed with two hidden layers of $d$ neurons with ELU activation. 
Each benchmark method is initialized and seeded identically with the same random weights.  For dataset preprocessing, all continuous variables are standardized with a mean of 0 and a variance of 1. 
We train each model using the Adam optimizer with a learning rate of 0.0005 for up to a maximum of 300 epochs.

\paragraph{Computational hardware.}  All models were trained on an Ubuntu 18.04 OS with 64GB of RAM (Intel Core i7-6850K CPU @ 3.60GHz) and 2 NVidia 1080 Ti GPUs.

\begin{figure}[!htbp]
\begin{center}
    \includegraphics[width=0.5\linewidth]{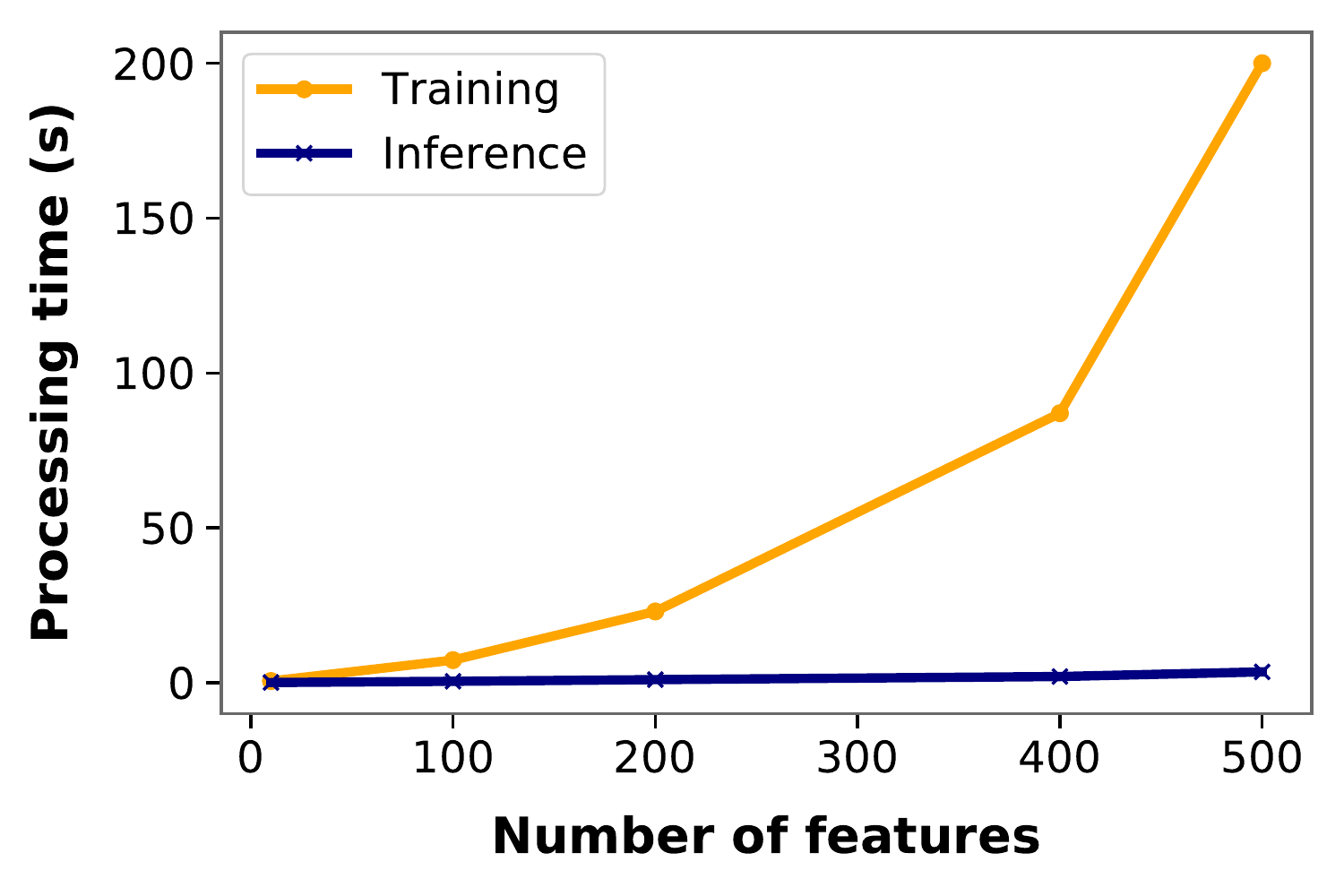}
  \end{center}
    \vspace{-10pt}
  \caption{\textbf{MIRACLE scalability analysis}}
\label{fig:complexity}
\end{figure}

\section{Computational Complexity}\label{appdx:complexity}

\begin{algorithm}[!htbp]
   \caption{Train MIRACLE}
   \label{alg:miracle}
\begin{algorithmic}
   \STATE {\bfseries Input:} An incomplete dataset $\mathbf X$ with missing values, a missing indicator matrix $\mathbf R$ (with 1 indicating observed), an imputed matrix $\tilde{\mathbf X}^{(0)}$ by some baseline method,  
    \STATE {\bfseries Output:} Imputed dataset $\tilde{\mathbf X}^{*}$ with no missing values.
   \STATE \textbf{Initialization:} Imputation network $f $, $\mathcal{G} = \varnothing$  with maximum size $M_\mathcal{G}$ for saving imputed matrices over epochs
   \REPEAT
    \STATE Train $f$ for one epoch by optimizing the objective function $\mathcal{L} = \mathcal{L}_1 + \beta_1\mathcal{R}_1 +  \beta_2\mathcal{R}_2$ with $\tilde{\mathbf X}^{(0)}$ as input.
    \IF {$\mathcal{G} $ is full}
        \STATE Remove the first element from $\mathcal{G} $
    \ENDIF
    \STATE $\tilde{\mathbf X}\leftarrow f (\tilde{\mathbf X}^{(0)}) $,   $\mathcal{G} \leftarrow \mathcal{G}\cup \{ \tilde{\mathbf X}\}$  
    \STATE $\tilde{\mathbf X}^{(0)} \leftarrow$ average all the elements of $\mathcal{G}$.
   \UNTIL{MIRACLE converges (i.e., change of $\mathbf X^{(0)} $ is small) }
   \STATE \textbf{return} $\tilde{\mathbf X}^{*}\leftarrow \tilde{\mathbf X}^{(0)}$
\end{algorithmic}
\end{algorithm}
\vspace{-.40cm}

Pseudocode for MIRACLE is provided in Algorithm~\ref{alg:miracle}.
We perform an analysis of the MIRACLE scalability.  Using our synthetic data generation, we created datasets of 1000 samples.  Using our the synthetic experimental setup presented in the main paper, we present the computational timing results for MIRACLE as we increase the number of input features on inference and training time in Figure~\ref{fig:complexity}.  Computational time scales linearly with increasing the number of input samples.  As expected, we observe that the time to train 1000 samples grows exponentially with the feature size; however, the inference time remains linear.  Inference time on 1000 samples with 400 features takes approximately 1.1 seconds, while training time takes nearly 85 seconds.  Experiments were conducted on an Ubuntu 18.04 OS using 6 Intel i7-6850K CPUs.

\section{Ablation study}\label{appx:ablation}

We provide an ablation study on our MIRACLE loss function in Eq.~\ref{eq:overall_loss} to understand the sources of gain of MIRACLE.  Here we execute this experiment on our real datasets using the same experimental details highlighted in the main manuscript.  We show the results of our ablation on MIRACLE using MissForest as baseline imputation with MAR missingness to highlight our sources of gain in Table~\ref{table:ablation}.  Here, we observe that MIRACLE  (rightmost column) has the most gain over all datasets.  Additionally, we observe that $\mathcal{L}_1 + \mathcal{R}_1 + \mathcal{R}_2$
has the most gain when MIRACLE has the most performance improvement over the baseline (see Fig.~\ref{fig:improvements} in the manuscript).

\begin{table*}[!htbp]
\caption{\textbf{Ablation study of MIRACLE on real datasets to highlight sources of gain.}}\label{table:ablation}
\setlength\tabcolsep{6pt}
\begin{center}
\begin{tabular}{lcccc }
 \toprule
Dataset& $\mathcal{R}_1 + \mathcal{R}_2$ & $\mathcal{L}_1 + \mathcal{R}_2$ & $\mathcal{L}_1 + \mathcal{R}_1$  & $\mathcal{L}_1 + \mathcal{R}_1 + \mathcal{R}_2$  \\
 \midrule
abalone & 0.321 $\pm$ 0.108 & 0.521 $\pm$ 0.199 & 0.312 $\pm$ 0.082 & 0.222 $\pm$ 0.062  \\
autism & 0.093 $\pm$  0.005 & 0.094 $\pm$ 0.004 & 0.091 $\pm$ 0.004 & 0.073 $\pm$ 0.004 \\
energy & 0.106 $\pm$ 0.011 & 0.147 $\pm$ 0.077 & 0.132 $\pm$ 0.050 & 0.065 $\pm$ 0.061  \\
protein &  0.134 $\pm$ 0.016 & 0.129 $\pm$ 0.008 & 0.119 $\pm$ 0.010 & 0.080 $\pm$ 0.008\\
life expectancy & 0.239 $\pm$ 0.007 & 0.223 $\pm$ 0.019 & 0.216 $\pm$ 0.014 & 0.208 $\pm$ 0.015  \\
community & 0.490 $\pm$ 0.015 & 0.516 $\pm$ 0.020 & 0.479 $\pm$ 0.023 & 0.463 $\pm$ 0.010  \\
yeast & 0.984 $\pm$ 0.013 & 0.984 $\pm$ 0.006 & 0.988 $\pm$ 0.004 & 0.950 $\pm$ 0.014  \\
mammo masses & 1.105 $\pm$ 0.010 & 1.150 $\pm$ 0.009 & 1.103 $\pm$ 0.013 & 1.040 $\pm$ 0.013  \\
wine quality & 0.797 $\pm$ 0.004 & 0.745 $\pm$ 0.013 & 0.724 $\pm$ 0.008 & 0.722 $\pm$ 0.003  \\
facebook & 1.056 $\pm$ 0.005 & 1.032 $\pm$ 0.044 & 1.034 $\pm$ 0.056 & 0.983 $\pm$ 0.002 \\

\bottomrule
\end{tabular}
\end{center}
\end{table*}

\section{Understanding missingness location} \label{apx:missingness}

\begin{figure}[H]
    \centering
    \includegraphics[width=0.5\linewidth]{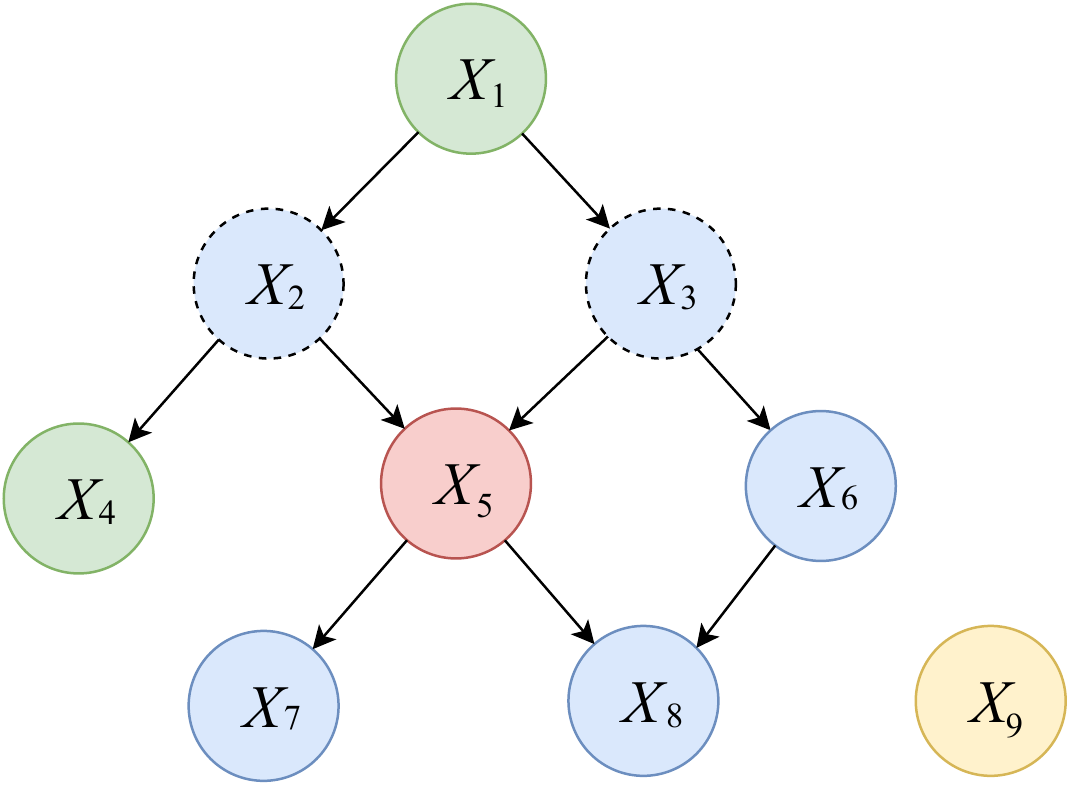}
    \vspace{-0.0cm}
    \caption{\textbf{A sample DAG.}  $X_5$ is the incomplete variable in red.  The Markov Blanket \texttt{MB}$(X_5)$ is shown in blue, and the causal parents \text{Pa}$(X_5)$ are shown with dashed borders.  $X_9$ represents a variable that is not causally linked to anything.
    }
    \vspace{-0.5cm}
    \label{fig:sampleDAG}
\end{figure}

An important consideration is how well does predicting with the causal parents work when down-selecting features.  Consider missingness in $X_5$ in the DAG in Figure~\ref{fig:sampleDAG}. The first column with the causal parents \text{Pa}$(X_5)$ mean that only the parents of features were used for imputation. $X_9$ represents a variable that is not causally linked to anything. 

Using our synthetic data generating process, we synthesized a dataset according to Figure~\ref{fig:sampleDAG}.  
The goal here was to impute the missing values in $X_5$, using each variable in Figure~\ref{fig:sampleDAG} to induce the missingness. 
Each of the missingness causes is categorized as MAR, except for $X_5$, which is MNAR (since missingness caused by itself), and for $X_9$, which is MCAR, since it is an external noise variable. The results provide several interesting findings.
\begin{enumerate}[leftmargin=*,itemsep=0pt]
    \item Using MissForest as a baseline imputer, the results in Table \ref{table:synth_location} show that MIRACLE performs as well as \text{Pa}$(X_5)$, and has better performance than the baseline imputer. Moreover, the two right-most columns of Table \ref{table:synth_location} give the average estimated functional dependence of $X_5$ (our target for prediction) and its parents and non-parents. We see that MIRACLE recovers true parents consistently.
    \item We see that using causal parents (\text{Pa}$(X_5)$ and MIRACLE) for missingness caused by itself, $X_5\ddagger$, and a noise variable, $X_9\dagger$, leads to the least amount of improvement.
    \item We see that MIRACLE has the most gain when the missingness is caused by a causal parent ($X_2$ or $X_3$).
    \item Interestingly, for this example, we observe comparable performance when using the Markov blanket features versus all features in our baseline algorithm (MissForest).  This suggests that the Markov blanket features are likely used for imputation by the baseline method.
\end{enumerate}

In Appendix~\ref{appdx:additionalsynth}, we show that the baseline performs similarly to the Markov blanket features colored in blue in Figure~\ref{fig:sampleDAG}.

\begin{table}[!ht]
    \vspace{-0.20cm}
\caption{\textbf{Understanding the location of missingness}. We predict $X_5$ when its missingness is caused by each variable in the DAG. $\ddagger$ and $\dagger$ represent MNAR and MCAR, respectively.  All other causes are MAR. The two right-most columns show the learned edge weights into $X_5$ for the parental and non-parental variables.}\label{table:synth_location}
\vspace{-0.0cm}
\begin{center}
\resizebox{0.9\linewidth}{!}{
\setlength\tabcolsep{6pt}
\begin{tabular}{l|cccc|cc}
 \toprule
  & \multicolumn{4}{c|}{$X_5$ imputed error (RMSE)} & \multicolumn{2}{c}{$X_5$ edge weights (no threshold)} \\
\cmidrule(lr){2-5} \cmidrule(lr){6-7}
 
 Cause & \text{Pa}$(X_5)$ & \texttt{MB}$(X_5)$ & Baseline & MIRACLE & Pa & non-Pa \\
 \midrule
$X_1$ & 0.11 $\pm$ .06 & 0.15 $\pm$.03 & 0.27 $\pm$ .05 & \textbf{0.12 $\pm$ .07} & 0.44 $\pm$ 0.14 & 0.02 $\pm$ 0.01\\
$X_2$ & 0.98 $\pm$ .08& 1.34 $\pm$ .05 & 1.31 $\pm$ .06 & \textbf{0.49 $\pm$ .06} & 0.64 $\pm$ 0.09 & 0.01 $\pm$ 0.01\\
$X_3$ & 1.20 $\pm$ .04 & 1.49 $\pm$.04 & 1.45 $\pm$ .09 & \textbf{0.50 $\pm$ .06} & 0.62 $\pm$ 0.13 & 0.02 $\pm$ 0.01\\
$X_4$ & \textbf{0.69 $\pm$ .05} & 1.20 $\pm$.07 & 1.23 $\pm$ .05 & 1.04 $\pm$ .05 & 0.29 $\pm$ 0.11 & 0.13 $\pm$ 0.05\\
$X_5\ddagger$& \textbf{1.51 $\pm$ .03} & 1.75 $\pm$.08 & 1.76 $\pm$ .06 & 1.59 $\pm$ .07 & 0.37 $\pm$ 0.18 & 0.03 $\pm$ 0.02\\
$X_6$ & \textbf{0.13 $\pm$ .08 }& 0.17 $\pm$.04 & 0.18 $\pm$ .07 & 0.14 $\pm$ .05 & 0.34 $\pm$ 0.15 & 0.05 $\pm$ 0.02\\
$X_7$ & 1.04 $\pm$ .05 & 1.47 $\pm$.04 & 1.47 $\pm$ .06 & \textbf{1.01 $\pm$ .06} & 0.39 $\pm$ 0.05 & 0.04 $\pm$ 0.01\\
$X_8$ & 0.21 $\pm$ .04 & 0.28 $\pm$.05 & 0.23 $\pm$ .03 & \textbf{0.20 $\pm$ .03} & 0.46 $\pm$ 0.10 & 0.02 $\pm$ 0.01\\
$X_9\dagger$ & 0.15 $\pm$ .03 & 0.18 $\pm$.04 & 0.17 $\pm$ .07 & \textbf{0.14 $\pm$ .05} & 0.31 $\pm$ 0.15 & 0.02 $\pm$ 0.01\\

\bottomrule
\end{tabular}
}
\vspace{-0.6cm}
\end{center}
\end{table}

\section{Convergence}\label{appx:convergence}

Using the same experimental setup, in Appendix~\ref{apx:missingness}, we examine how the estimated adjacency matrix converges to the truth as sample size increases.  Table~\ref{table:convergence}, shows that as sample size increases we see that the quality of the learned m-graph improves.

\begin{table}[!ht]
    \vspace{-0.20cm}
\caption{\textbf{Convergence of DAG weights as dataset size increases.} $\circ$ is the element-wise product and $\hat{W}$ is the predicted adjacency matrix weights}\label{table:convergence}
\vspace{-0.0cm}
\begin{center}

\begin{tabular}{l|c}
 \toprule
Dataset size & $\sum_{i,j}{(\hat{W} \circ W)}/ \sum_{i,j}{\hat{W}}$ \\
\midrule

100       &       0.135 $\pm$ 0.04 \\
500        &             0.242  $\pm$0.03 \\
1000    &              0.252  $\pm$0.03 \\
5000        &           0.910  $\pm$0.01\\
10000        & 0.916  $\pm$0.01 \\
 \bottomrule
\end{tabular}

\vspace{-0.6cm}
\end{center}
\end{table}

\end{document}